\documentclass[lettersize,journal]{IEEEtran}
\usepackage{amsmath,amsfonts}
\usepackage{algorithmic}
\usepackage{algorithm}
\usepackage{array}
\usepackage[caption=false,font=normalsize,labelfont=sf,textfont=sf]{subfig}
\usepackage{textcomp}
\usepackage{stfloats}
\usepackage{url}
\usepackage{graphicx}
\usepackage{cite}

\usepackage{multirow}
\usepackage{pifont}
\newcommand{\cmark}{\ding{51}}%
\newcommand{\xmark}{\ding{55}}%
\usepackage{float}
\usepackage{booktabs} 
\usepackage{soul}

\begin{document}

\title{Mean Height Aided Post-Processing \\for Pedestrian Detection}

\author{Jing Yuan, Tania Stathaki,~\IEEEmembership{Member, IEEE}, Guangyu Ren

\thanks{J. Yuan and T. Stathaki are with the Department of Electrical and Electronic Engineering, Imperial College London, London, SW7 2AZ UK (e-mail:j.yuan20@imperial.ac.uk, t.stathaki@imperial.ac.uk)}
\thanks{G. Ren is with the Department of Computer Science, University College London, r.guangyu@ucl.ac.uk.}}

\maketitle

\begin{abstract}
The design of pedestrian detectors seldom considers the unique characteristics of this task and usually follows the common strategies for general object detection.
To explore the potential of these characteristics, we take the perspective effect in pedestrian datasets as an example and propose the mean height aided suppression for post-processing.
This method rejects predictions that fall at levels with a low possibility of containing any pedestrians or that have an abnormal height compared to the average.
To achieve this, the existence score and mean height generators are proposed.
Comprehensive experiments on various datasets and detectors are performed; the choice of hyper-parameters is discussed in depth.
The proposed method is easy to implement and is plug-and-play.
Results show that the proposed methods significantly improve detection accuracy when applied to different existing pedestrian detectors and datasets. 
The combination of mean height aided suppression with particular detectors outperforms state-of-the-art pedestrian detectors on Caltech and Citypersons datasets.
\end{abstract}

\section{Introduction}
\label{sec: introduction}
\IEEEPARstart{P}{edestrian} detection aims to locate and identify persons from an image.
It serves as the basis of various computer vision tasks \cite{RN62}, such as person re-identification \cite{RN197, RN198}, person search \cite{RN199}, and human pose estimation \cite{RN201}.
It also plays an important role in autonomous driving \cite{RN203, RN204} and video surveillance \cite{RN205}.
As a downstream task of general object detection, it follows the typical pipeline:
The input image is fed into a detector to extract features.
These features are sent to the classifier and regressor to generate the confidence scores and bounding boxes of potential pedestrians.
Then, in the post-processing phase, redundant predicted pedestrians are reduced by Non-Maximum Suppression (NMS), producing the final detection results.
The performance of pedestrian detection is commonly evaluated with the log-average miss rate.
Traditional detectors extract handcrafted features such as Histogram of Oriented Gradients (HOG) \cite{dalal2005hog}, which contains the edge information in all directions and Aggregated Channel Features (ACF) \cite{dollar2014acf}, which concatenates the silhouette feature and colour feature.
In recent years, Deep Neural Networks (DNN) based detectors have been proposed to automatically extract more expressive semantic features by optimizing detector parameters on a training set.
As DNN detectors dramatically outperform traditional ones, they have been widely used in object detection tasks.

Detecting pedestrians is challenging due to their varying sizes and frequent obstruction, resulting in a decrease in the availability of valuable target characteristics.
To improve the detection performance further, pedestrian detectors borrow DNN structures developed for general object detection, such as methods proposed in \cite{cascade-rcnn, faster-r-cnn}.
Most detectors focus on extracting better features through the network \cite{liu2019CSP, cascade-rcnn}, optimizing predictions progressively \cite{ALF, cascade-rcnn} etc.
Since pedestrians are an important category within general objects, they can seamlessly integrate into the workflow and feature extraction approach employed for detecting such objects.
And those general object detectors work effectively for pedestrian detection. In this case, pedestrians are gradually treated the same as other categories. However, this results in relatively under-explored unique characteristics of pedestrian datasets.

\begin{figure}[!t]
\centering
\subfloat[]{\includegraphics[width=0.1\textwidth]{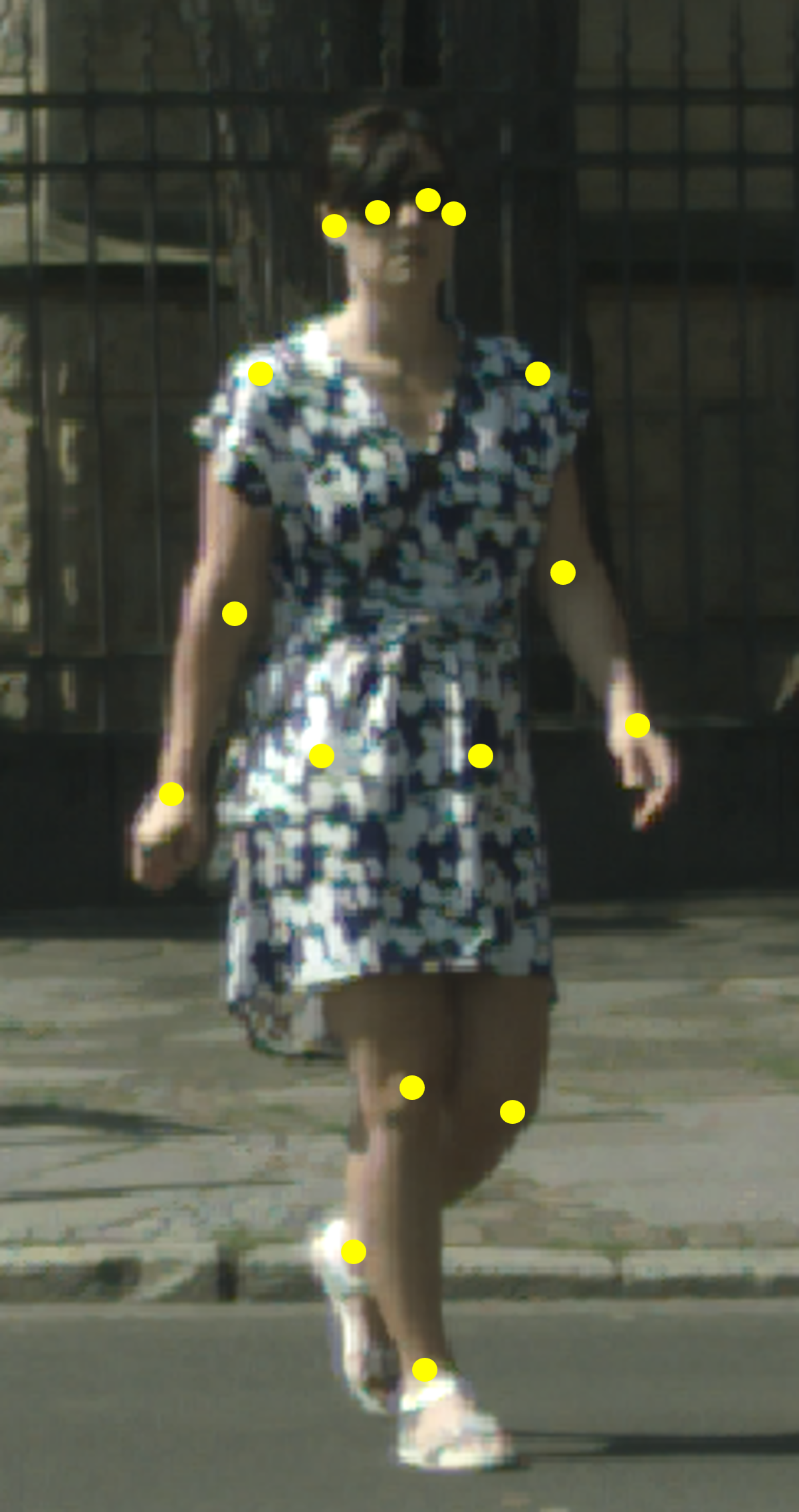}%
}
\hfil
\subfloat[]{\includegraphics[width=0.1\textwidth]{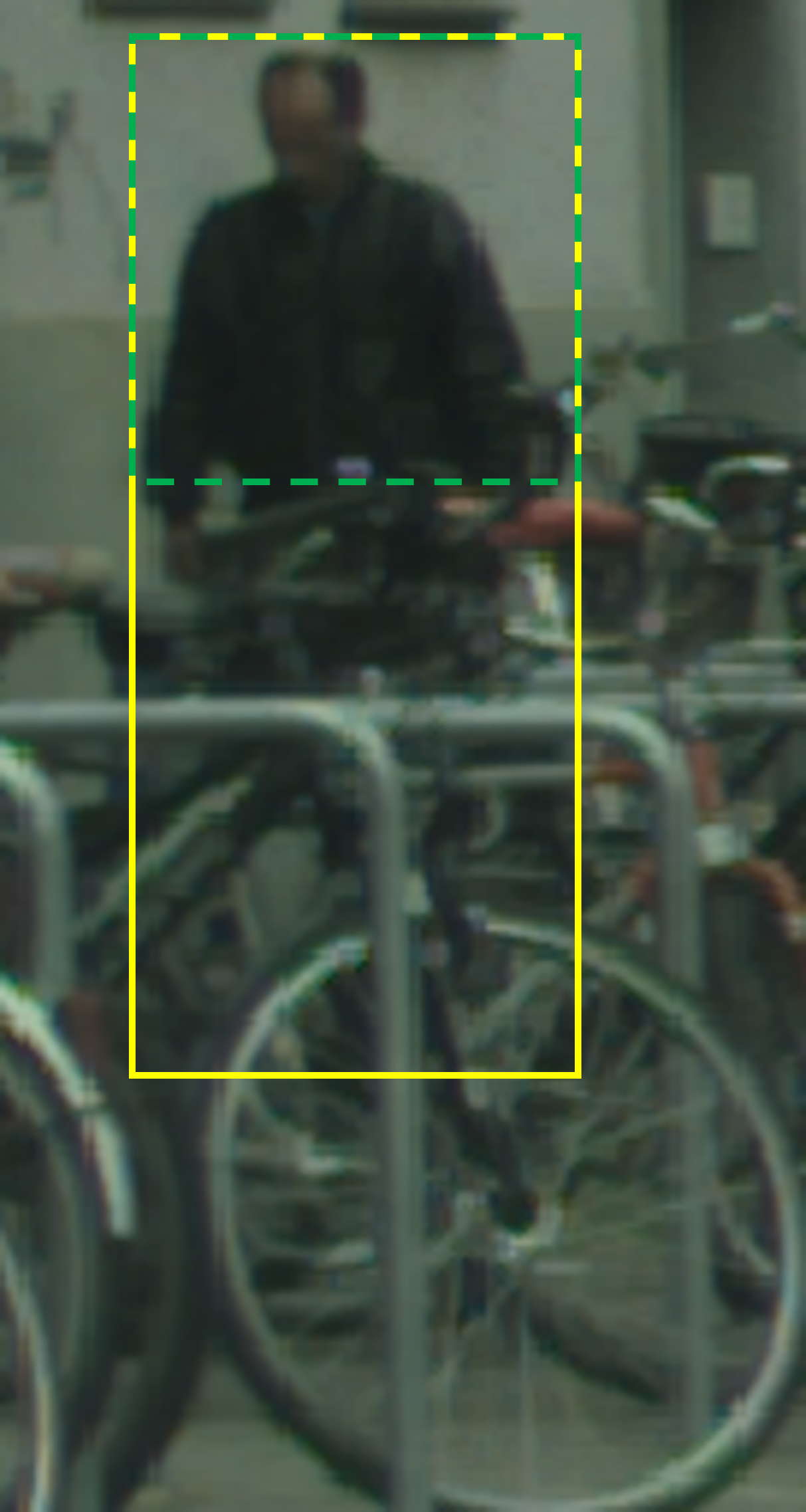}%
}
\caption{Two examples of unique characteristics of a person and pedestrian dataset annotations. (a) Key points of a person used in \protect\cite{zhang2020kgsnet}. (b) Visible part (green dashed line) and complete body (yellow line) annotations used in \protect\cite{chi2020pedhunter, mgan2019, MGAN+}.}
\label{fig: unique characteristics}
\end{figure}

Pedestrians have unique characteristics.
They are usually upright human bodies with similar aspect ratios.
They have recognizable silhouettes forming the shape of the bodies.
Human body parts, namely the head, arms and legs, appear at fixed locations.
Some pedestrian datasets even provide annotations of visible parts.
These characteristics can also help to improve detection accuracy. 
For example, \cite{zhang2020kgsnet} utilizes a key point detector to distinguish the body parts, such as eyes, ears, shoulders, elbows, hands, legs, etc., as depicted in Fig. \ref{fig: unique characteristics}(a), for the following classification.
\cite{mgan2019, MGAN+, chi2020pedhunter} predict the visible parts, Fig. \ref{fig: unique characteristics}(b) right, of pedestrians.
These methods implicitly guide the feature extraction of the neural networks or explicitly enhance the contribution of visible features and suppress the interruption of occluded parts.
Thus, with such unique information, pedestrian detection performance is improved significantly. 
Note that the body structure information requires high-resolution pedestrians while the visible area requires extra labelling time besides labelling full boxes.

\begin{figure}[!t]
\centering
\subfloat[]{\includegraphics[width=0.2\textwidth]{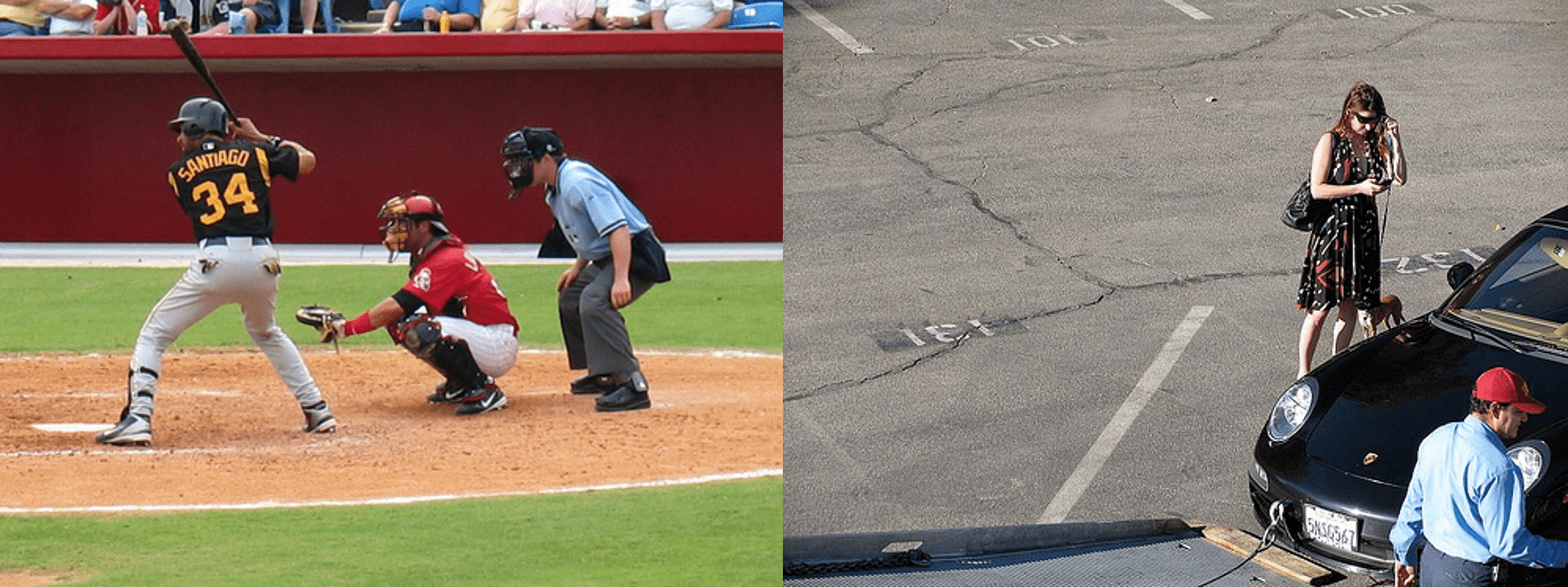}%
}
\hfil
\subfloat[]{\includegraphics[width=0.2\textwidth]{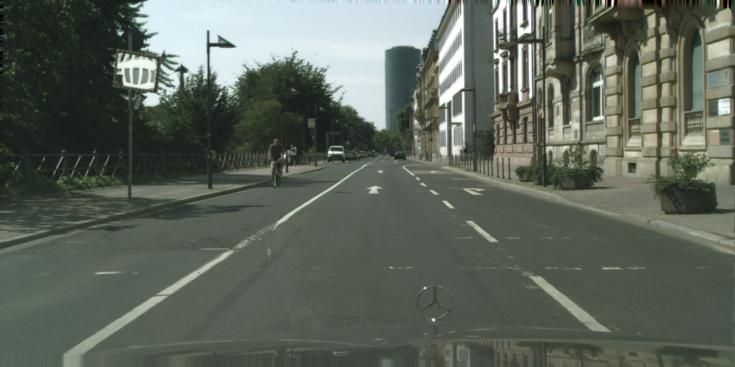}%
}
\caption{Our motivation relies on the observation that pedestrian datasets possess the unique characteristic of perspective effect compared to general object datasets.
(a) Two sample images containing persons from COCO dataset \protect\cite{coco}. 
A weak perspective effect is observed. The inclusion of the person category demonstrates the feasibility of applying general detectors to pedestrian detection tasks. 
(b) One typical sample image from Citypersons dataset \protect\cite{zhang2017citypersons}. 
A severe perspective effect is observed with parallel lane lines converging at one point.}
\label{fig: dataset samples}
\end{figure}

\begin{figure}[!t]
\centering
\includegraphics[width=2.0in]{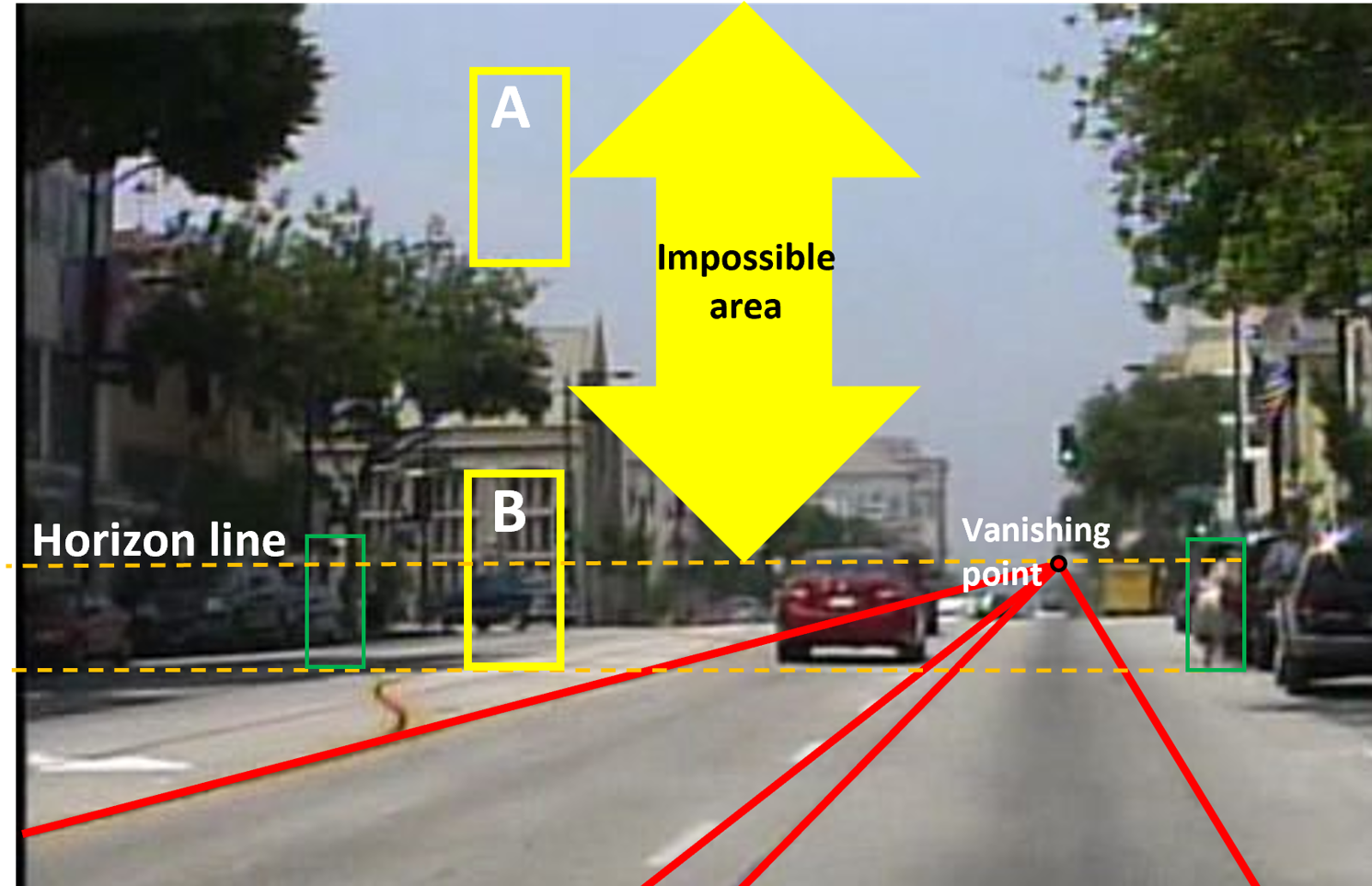}%
\caption{Our proposed method can be better understood by examining an image from the Caltech pedestrian dataset \protect\cite{dollar2011pedestrian} that has a severe perspective effect.
Parallel lane lines are marked in red and extended to converge at the vanishing point marked with a black circle.
The higher dashed line is the horizon line.
The two principles that underlie this method can be illustrated through two sample bounding boxes.
Yellow box A stands in the impossible area over the horizon line.
Objects in this area mean they are standing above the ground plane, which is impossible for pedestrians in this image.
Yellow box B stands at the same level (lower dashed line) as two ground truths (green boxes) but is too large for a normal pedestrian.}
\label{fig: perspective effect}
\end{figure}

In this paper, we explore the widely existing perspective effect in pedestrian datasets and investigate the potential of this easily available unique characteristic in improving detection accuracy.
Unlike the general object dataset (Fig. \ref{fig: dataset samples}(a)), parallel lines in pedestrian datasets are observed to converge at a point known as the vanishing point, as shown in Fig. \ref{fig: dataset samples}(b).
Commonly used pedestrian datasets, such as Citypersons dataset \cite{zhang2017citypersons} and Caltech dataset\cite{dollar2011pedestrian}, are frames of driving video record sequences taken by the camera set on a vehicle, usually a car.
Therefore, the frames within the same sequence maintain consistent eye levels and have noticeable perspective effects.
Within such images, pedestrians exhibit a linear reduction in size, eventually reaching infinitesimally small proportions at the horizon line, which is composed of vanishing points (as depicted in Fig. \ref{fig: perspective effect}).
Although severe perspective distortion introduces significant variations in pedestrian sizes, thereby hindering detection, we can still leverage this effect to our advantage.
To elaborate further, we establish two key principles to guide our approach.
Firstly, no pedestrians should be positioned above the horizon line, as this would imply an impossible scenario where pedestrians exist in the sky.
Second, the height of an average person should not deviate excessively from the mean height of pedestrians at their respective levels.
Here, 'level' refers to the distance between the target and the camera plane, wherein pedestrians sharing the same distance are situated on the same level.
This alignment manifests as pedestrians having their feet aligned within the same row of pixels in an image.
By adhering to these principles, we can identify and reject bounding box A (Fig. \ref{fig: perspective effect}), which is situated above the horizon line, indicating a pedestrian in the sky.
Similarly, bounding box B (Fig. \ref{fig: perspective effect}) should be rejected as it exceeds the typical height of ordinary pedestrians, as denoted by the green boxes.
To address these challenges, we introduce the Mean Height Aided Suppression (MHAS) method, which follows the aforementioned rules to suppress bounding boxes that do not adhere to these fundamental principles.

Compared to the unique body structures and visible part annotations, the mean height information can be obtained easily with existing full body annotations, saving additional labelling time. 
In addition, it widely exists in both high- and low- resolution images, while sometimes the body structures are too small to be recognized.
MHAS improves the performance of existing pedestrian detectors, especially single-stage ones.
The combination of MHAS with specific detectors achieves the best performance among the state-of-the-art.
This method is designed to be plug-and-play, making it easy to use.
Comprehensive experiments show that the proposed method is robust on various datasets and detectors.
The choice of hyper-parameters in MHAS is investigated in depth.

Our main contributions are as follows:
\begin{itemize}
    \item Propose the convenient Mean Height Aided Suppression (MHAS) in the post-processing stage to improve detection accuracy across various detectors and datasets.
    \item Demonstrate the potential of under-explored unique characteristics in pedestrian datasets, namely the mean height, which might inspire the design of scene-specific detectors.
\end{itemize}

\section{Related Work}
\label{sec: related work}
\subsection{Common Pedestrian Detectors}
Most pedestrian detectors also possess the capability for general object detection.
For instance, some detectors employ techniques such as fusing multi-scale features to capture detailed information and high-level semantic context, thereby enabling the detection of objects with varying scales, including pedestrians.
\cite{cascade-rcnn} utilizes the feature pyramid network to generate multi-scale feature maps from up to down. 
These upsampled features contain high-level information, compensating for the lack of representation in lower features.
\cite{csp+vit} uses deformable vision transformers to adaptively combine features across scales at learned sampling locations with learned attention weights.
Such methods enhance the representation of features, so their ideas or structures can also be used to detect other categories.

Some detectors \cite{ALF, cascade-rcnn} make predictions progressively and repeatedly with fused or separate feature maps.
\cite{ALF} detects varying-scale pedestrians with multiple branches. 
Each branch takes a different scale feature as input and refines the bounding boxes through two-round predictions.
\cite{cascade-rcnn} predicts bounding boxes, then extracts features, which are more focused on targets, in this region and sends them to the next round of prediction repeatedly.
To facilitate NMS, \cite{reploss} proposes the repulsion loss to predict more compact bounding boxes for crowds under significant occlusion.

These algorithms are effective and can be widely applied to various object detection tasks besides pedestrian detection.
In this case, people may question the difference between pedestrian detectors and general object detectors, which have been relatively under-explored so far.

\subsection{Pedestrian Specific Detectors}
Some detectors utilize the unique characteristics of pedestrians to achieve further improvements.
For example, \cite{zhang2020kgsnet} takes advantage of the body structures.
It applies a key-point detector to each proposal region to locate the key body parts, like the head, shoulders, elbows, arms, hands, legs and feet, as sketched in Fig. \ref{fig: unique characteristics}(a). 
These semantic body parts are padded to the original proposal at the image level, leading to more details and structural information.
It is worth noting that for accurate body part identification, relatively small proposal regions are upsampled using a super-resolution network before estimating body parts.
\cite{mgan2019, MGAN+} and \cite{chi2020pedhunter} train masks of visible parts and heads to guide detection implicitly or explicitly. 
\cite{mgan2019, MGAN+} predict a visible mask of a proposal region to enhance the features of the visible body parts and reduce the interruption of background information.
\cite{chi2020pedhunter} removes the mask prediction in the inference stage.
In these methods, additional mask annotations of the human head and visible parts are required, which is time-consuming.
\cite{liu2019CSP} only predicts the height of the pedestrians and uses the constant aspect ratio of pedestrian bounding boxes to compute the corresponding width.

\subsection{Perspective Application in Object Centric Tasks}
In object tracking, \cite{leibe2008coupled} implements scene geometry estimation in tracking from static cameras in a moving vehicle.
The system makes use of the ground plane in the scene to narrow down potential object locations. 
It also incorporates camera calibration to combine separate detections into trajectories within a world coordinate frame over a set period.
Perspective also plays an important role in 3D object detection.
\cite{hoiem2008putting} shows that probabilistic estimates of 3D geometry can accurately place objects in perspective and help model scale and location variations in images, aiding the understanding of the interplay among various elements.
The concept of perspective commonly emerges as the projection between 2D and 3D domains.
For example, \cite{chen2016monocular} improves 3D object detection in autonomous driving by sampling bounding boxes based on ground plane size assumptions and projecting them onto the image plane, eliminating the requirement for laborious multi-scale searching.
\cite{liang2018MultiSensor3D} projects the image features into Bird's Eye View (BEV) and fuses projected features with the BEV LiDAR feature maps for multi-sensor 3D object detection.
\cite{huang2019perspectivenet} utilizes 2D perspective points as a means to connect the 2D image plane and the 3D world. 
These points are the 2D projections of the 3D bounding box corners and provide valuable 3D information like position and orientation.
\cite{kocur2020perspectiveTransformation} involves using the geometry of vanishing points in the scene being monitored to create a perspective transformation.
Such transformation allows for a more intuitive approach to detecting bounding boxes by simplifying the process from 3D to 2D.
\cite{wang2022probabilistic} model the geometric relations across different objects to enhance depth estimation from contextual connections for monocular 3D object detection.
\section{Method}
\label{method}
\begin{figure*}[!t]
\centering
\includegraphics[width=\textwidth]{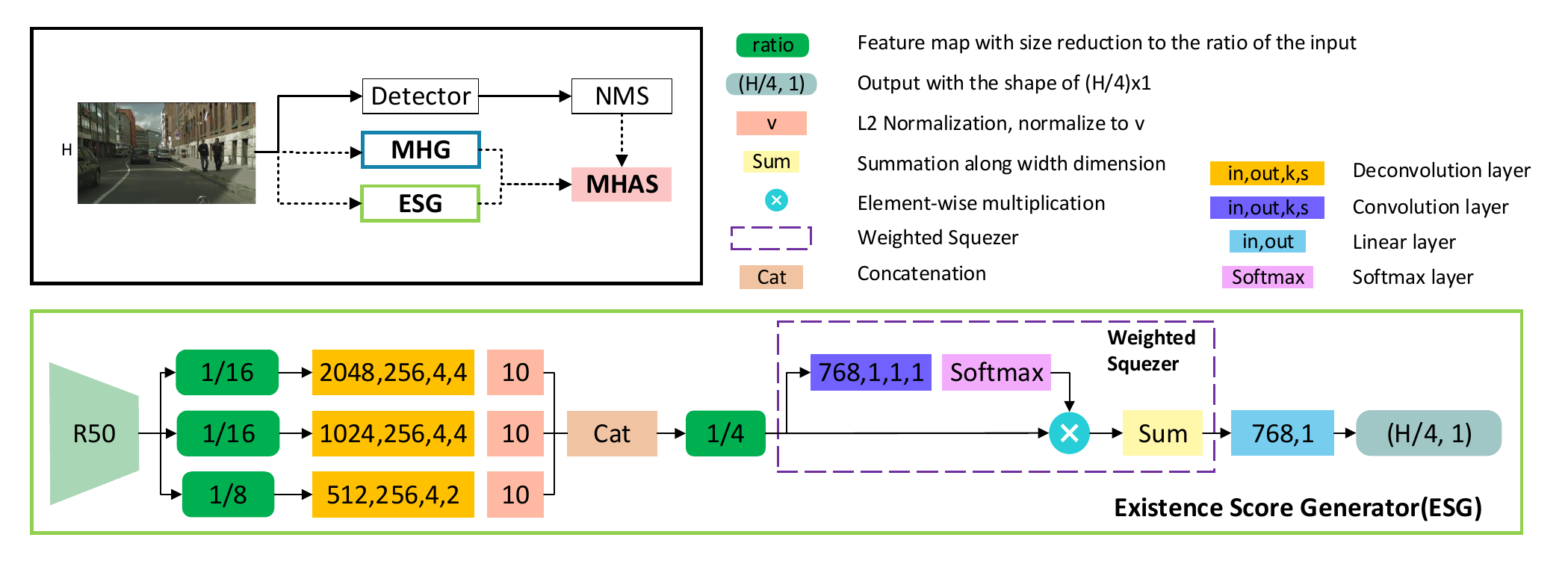}%
\caption{The black box shows the workflow of pedestrian detection with the proposed Mean Height Aided Suppression (MHAS) method.
The greed box shows the architecture of the embedded Existence Score Generator (ESG) block.
In the workflow, solid lines depict the steps in conventional pedestrian detection, while dashed lines illustrate the proposed components.
The term 'Detector' refers to any pedestrian detection algorithm employed.
'MHG' stands for Mean Height Generator.
Details of MHG and MHAS are introduced in Section \ref{sec: mhg} and \ref{sec: mhas}.
}
\label{fig: mhas-workflow}
\end{figure*}

Typically, in the conventional pipeline, an image is fed into the pedestrian detector, which generates confidence scores and corresponding bounding boxes. 
Subsequently, these output results are passed through a Non-Maximum Suppression (NMS) algorithm.
With the use of Mean Height Aided Suppression (MHAS), the workflow is presented as Fig. \ref{fig: mhas-workflow}.
Apart from normal steps, the input image is sent to an Existence Score Generator (ESG) module and a Mean Height Generator (MHG) module to produce the existence scores and mean heights, respectively.
The output of this predictor and NMS are fed into the MHAS module.
This module further suppresses the false positives and outputs the final confidence scores and bounding boxes.

We propose the MHAS method to suppress the false positives further.
This method requires the existence scores and (or) the mean heights. 
The complete method is the combination of both.
\subsection{Existence Score Generator (ESG)}
\label{sec: Existence Score Generator}
The existence score generator predicts the likelihood of a pedestrian being present at each level.
The higher the score, the more likely the corresponding level contains pedestrians' feet.
The overall architecture of the generator used in our experiments is illustrated in Fig. \ref{fig: mhas-workflow}.
This generator uses ResNet 50 \cite{he2016resnet} as the backbone. 
The features at the last three stages are upsampled and concatenated to form a high-resolution multi-layer feature map and sent to the Weighted Squeezer (WS).
A linear layer is used to predict the existence scores.

The key structure of the ESG is the Weighted Squeezer.
This module takes as input the concatenated multi-layer feature $\boldsymbol{x} \in \mathbb{R}^{B \times C \times H/4 \times W/4}$ where $B$, $C$, $H$ and $W$ refer to the batch size, number of channels, height and width of the input images after transformation. 
This feature map is sent to a convolution layer followed by a softmax layer along the width dimension to generate the weight $\boldsymbol{w}_{ws} \in \mathbb{R}^{B\times 1 \times H/4 \times W/4}$:
\begin{equation}
\label{eq:weighted squeeze-weights}
    \boldsymbol{w}_{ws} =  (Softmax(Conv(\boldsymbol{x})))
\end{equation}
which is then extended to $\boldsymbol{w}_{ws} \in \mathbb{R}^{B \times C \times H/4 \times W/4}$.
The output is the element-wise multiplication of weights and $\boldsymbol{x}$ and the summation along the width dimension:
\begin{equation}
\label{eq:weighted squeeze-elex}
    \boldsymbol{o}_{ws} = Sum(\boldsymbol{x} \cdot \boldsymbol{w}_{ws})
\end{equation}
where the output $\boldsymbol{o}_{ws} \in \mathbb{R}^{B\times C \times H/4}$ is invariant along the width dimension.
It is sent to two linear layers to predict the existence scores $\boldsymbol{s_e} \in \mathbb{R}^{B \times 1 \times H/4}$. 
The existence score is the possibility of having pedestrians' feet in the corresponding level. 

\begin{figure}[!t]
\centering
\includegraphics[width=0.8\columnwidth]{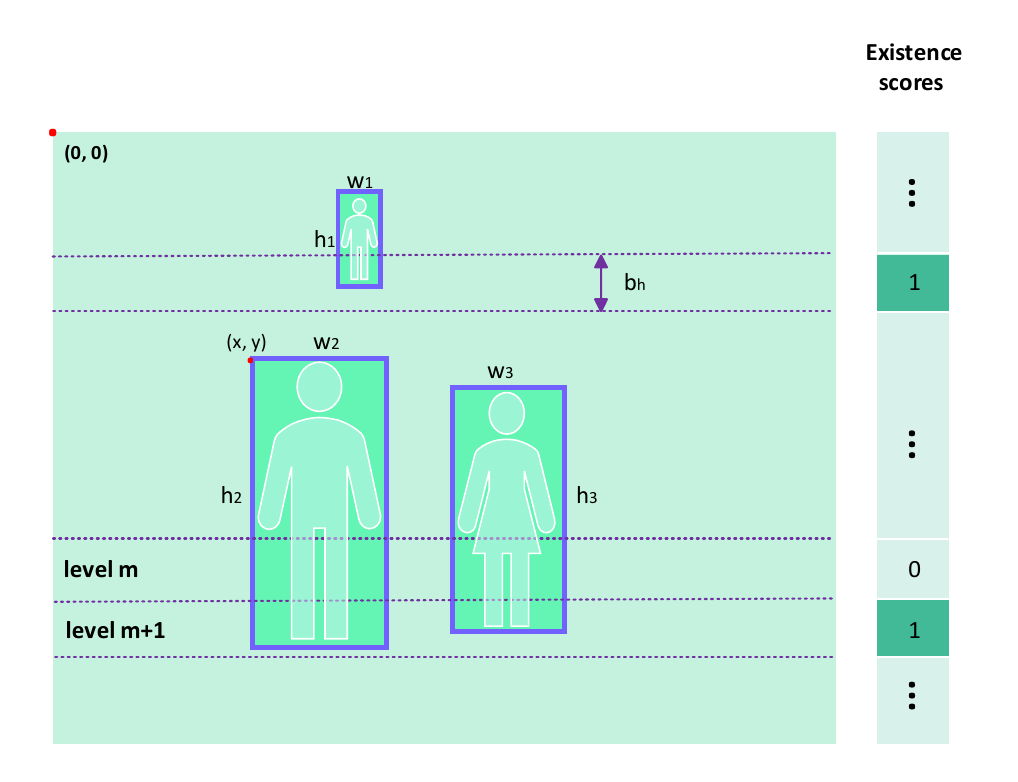}
\caption{The sketch of the ground truth existences scores and notations utilized in the training of the ESG module and the MHAS algorithm, respectively.
The left part shows an annotated image separated into non-overlapping levels.
Each level has a width of $b_h$ pixels.
The red point marks the coordinate origin of the images referenced in this paper.
The bounding box has a width of $w_i$ and a height of $h_i$ ($i=1,2,3$).
Its upper corner positions at $(x, y)$.
The right column is the ground truth existence score vector corresponding to each level.
The level that contains the bottom edge of the bounding box will be assigned as 1, otherwise, the value of 0 is set.}
\label{fig:MHAS-level-gt}
\end{figure}

An example of the ground truth existence score vector of an image is illustrated in Fig. \ref{fig:MHAS-level-gt}.
Each value in the vector corresponds to one level of the input image.
The score value is set as 0 at levels that contain pedestrian feet.
Values of other levels that contain no targets are set as 0 in both vectors.

The loss function $L_e$ is expressed as
\begin{equation}
L_e=-\frac{1}{K}\sum_{m=1}^{\frac{H}{4}}{\alpha_{m}\left(1-{\hat{p}}_{m}\right)^\gamma\log{\left({\hat{p}}_{m}\right)}}
\end{equation}

\begin{equation}
\begin{aligned}
{\hat{p}}_{m}=\left\{
\begin{aligned}
p_{m},&&y_{m}=1\\
1-p_{m},&&\mathrm{otherwise}
\end{aligned}
\right.
\end{aligned}
\end{equation}
\begin{equation}
\begin{aligned}
\alpha_{m}=\left\{
\begin{aligned}
1,&&y_{m}=1\\
\left(1-M_{m}\right)^\beta,&& \mathrm{otherwise}
\end{aligned}
\right.
\end{aligned}
\end{equation}
where $p_{m}\in\left(0,1\right)$ denotes the predicted existence score at $m$th level. The one-dimensional Gaussian filter $\boldsymbol{M}$ has variances proportional to the mean height at each one of the $K$ positive levels.
The hyper-parameters $\beta$, $\gamma$ and $\lambda$ are set as 4, 2 and 0.01 \cite{liu2019CSP}. 

\subsection{Mean Height Generator (MHG)}
\label{sec: mhg}
The mean height generator aims to produce the average height of pedestrians standing at each level.
The manual computation approach or perspective coefficient approach can be applied.
\subsubsection{Manual approach}
\label{sec: manual approach}
For simplification, the mean height can be calculated directly from the training set.
This relies on the assumption that the eye level of the whole dataset, including the training, validation, and test sets, are the same.
This approach leverages the extensive number of pedestrian samples within the dataset to compute the average height statistic of humans.
The accuracy of the manual mean height improves as the training set contains more pedestrians.
The more pedestrians the training set contains, the more accurate the manual mean height will be.

The pedestrians are annotated as $x, y, w, h$, representing the $x, y$ coordinates of the upper left corner of the bounding box and its width and height.
First, we divide the image into $H/4$ non-overlapping rows, each representing a level.
Then we average the heights with the corresponding bottom edges at the same level across the whole training set.
The final mean height is a vector of $H/4$ elements which will be shared for each image in MHAS.
The advantage of manual computation is that it is easy to apply, and no extra time is required in the inference stage.

However, one limitation of this approach is its reliance on a consistent eye-level assumption.
In cases where the eye level varies, the actual mean height at the same level can differ, leading to potential errors such as mistakenly rejecting true positives or failing to suppress false positives.
Besides, the more pedestrians the dataset has at each level, the more reasonable the mean heights are.

\subsubsection{Perspective coefficient approach}
\label{sec: perspective coefficient approach}
This is employed to overcome the issue of varying eye levels, which can occur when the camera position changes within the same video sequence or when the dataset comprises multiple video sources with varied eye levels. 
In such scenarios, the manual approach for calculating mean height becomes unreliable. 
To address this challenge, we adopt a predictive approach by estimating perspective coefficients. 
These coefficients allow us to generate adaptive mean heights for individual images, eliminating the need for assuming eye-level consistency. 
By leveraging perspective coefficients, we can account for variations in eye level and enhance the accuracy of pedestrian detection across diverse camera setups and video sources.

The perspective coefficients, namely $a$ and $b$, define the mapping from level to the mean height value with $\Bar{h}_m=a\cdot m + b$, where $m$ indexes the levels.
The linear perspective is applied as the average mean height ideally converges to a constant if the number of pedestrian samples is infinite and the images are taken on flat ground.

To reduce the inference time, a simple network is used to predict the perspective coefficients $a, b$.
It takes the image as input.
Like ESG, the backbone is ResNet 50, but only the feature map at the last stage, namely $\boldsymbol{x}$, is used.
The estimated coefficients vector $\boldsymbol{c}=[a, b]$ is
\begin{equation}
\label{eq: perspective ab}
    \boldsymbol{c} = Linear(g(f(\boldsymbol{x})))
\end{equation}
where $f$ is the L2 Normalization, which is also used in ESG as depicted in Fig. \ref{fig: mhas-workflow}, but each channel is normalized to 1 in this structure.
The function $g$ is 2D average pooling which outputs a vector feature. 
This is accomplished by adjusting the pooling kernel when the input image size changes.
The feature vector is fed into a straightforward linear regressor which predicts the two perspective coefficients $\boldsymbol{c}$.
Subsequently, the mean height at each level can be computed through linear perspective and the image-wise coefficients.

In the training stage, the smooth L1 loss is utilized to represent the distance between the predicted coefficients and the ground truth.
The ground truth $a, b$ for each image is generated by implementing linear regression.
To improve the accuracy of fitting, we have established the following rules for selecting the best images to form the training set.

(1) Images should have more than two pedestrian samples with bottom positions across more than two levels.
   This is because at least two samples are required for linear fitting.
   When crossing more levels, the fitting errors caused by box annotation inaccuracies and insufficient samples are relatively smaller.

(2) Images with relatively small $a$ (less than 0.1) are excluded from the training set.
   Small values of $a$ typically occur when samples have similar levels and heights, posing challenges for regression.
   This rule filters out unsuitable images that occasionally meet the first rule.

(3) The mean relative error between the heights calculated with regressed linear perspective and true heights should be kept within an acceptable range.
   This rule quantitatively describes the accuracy of the fitted parameters.
   Typically, images with smaller errors are considered more suitable for training when an adequate number of pedestrians are present.
   This rule is implemented to filter out images with pedestrians not standing on the same surface plane as the camera, for example, on an overpass or a slope.
   Such images may produce abnormal ground truths and confuse the predictor.

In summary, these rules ensure sufficient and relevant pedestrians in a single image for reasonable estimation of ground truth perspective coefficients.

\begin{figure}[!t]
\centering
\subfloat[]{\includegraphics[width=0.3\columnwidth]{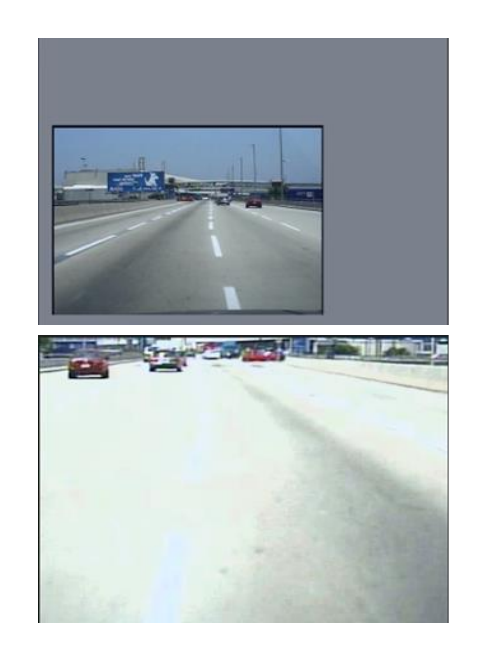}%
}
\hfil
\subfloat[]{\includegraphics[width=0.6\columnwidth]{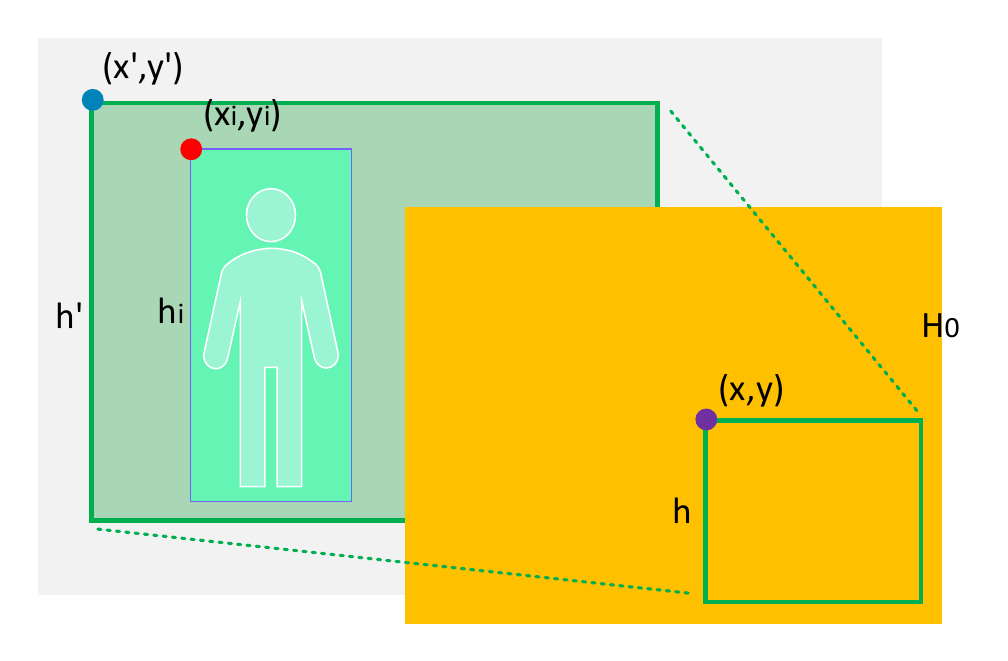}%
}
\caption{When using the perspective coefficient approach, the ground truths are affected by image transformations.
(a) Two examples of input images augmented by rescaling, cropping, and padding 
(b) Notations of the image transformations during data augmentation. 
The green bounded box is cropped from the original image (yellow) and rescaled to the image (green). 
This image is padded to the desired training size to produce the input image (grey).}
\label{fig:mhal-data augmentation}
\end{figure}

Considering image transformations, such as rescaling, random cropping and padding as shown in Fig. \ref{fig:mhal-data augmentation} (a), the ground truth coefficients should also be transformed to align with the corresponding transformations applied to the original image.
Suppose the original coefficients are $a_o, b_o$.
The area with upper left position $(x,y)$ and height $h$ is cropped as shown in Fig. \ref{fig:mhal-data augmentation} (b).
Then it is rescaled to the area with the height of $h'$ and padded to the training size with upper left position $(x',y')$ in the final image.
Note that $x, y, h$ used in this step do not represent bounding box annotations.
The corresponding ground truth $a, b$ are calculated as
\begin{equation}
    a = a_o    
\end{equation}

\begin{equation}
    b = \frac{h}{h'} \cdot b_o + a_o \cdot (y-y')
\end{equation}

The main advantage of this approach is that it is not limited by consistent eye levels, and minimal additional inference time is required.
The limitation is that there are relatively few qualified images and insufficient pedestrian samples, even within each selected image.
The former hinders training, while the latter decreases the accuracy of the ground truth. 
Note that for simplicity, the network structure is relatively straightforward.
A more sophisticated network or leveraging existing perspective estimation tools could potentially yield better coefficient estimation, although this aspect is not the primary focus of this paper.

\subsection{Mean Height Aided Suppression (MHAS)}
\label{sec: mhas}
The mean height information can be used in the post-processing stage to suppress the false positives after NMS.
The idea of MHAS is to reject predictions that fall at levels which are not likely to contain any targets or that heights deviate far from the average heights at corresponding levels.
The level is a band with a width of $b_h$ pixels, as shown in Fig. \ref{fig:MHAS-level-gt}. It is indexed by $m=1,2,..., \lceil H/b_h \rceil$ from top to bottom of the image.

Given a prediction with the bounding box $(x, y, w, h)$, the bottom edge falls at the level of $m_h=\lceil(y+h)/b_h\rceil$. 
It will be retained if the following conditions are satisfied at any level in the pre-designed range of $m \in [m_h-r, m_h+r]$:
\begin{equation}
    \label{eq:mh_filtering}
    |\frac{h}{\Bar{h}_m}-1|\leq \alpha_h
\end{equation}
\begin{equation}
    \label{eq:existence_score_filtering}
    s^m_e\geq \alpha_s
\end{equation}
where $\alpha_h$ and $\alpha_s$ are the thresholds for the mean height condition (Equation \ref{eq:mh_filtering}) and the existence score condition (Equation \ref{eq:existence_score_filtering}). 
$\Bar{h}_m(\Bar{h}_m \in \boldsymbol{\Bar{h}})$ is the mean height at $m$th level. $s^m_e(s^m_e \in \boldsymbol{s_e})$ is the existence score at $m$th level.
$\boldsymbol{\Bar{h}}$ and $\boldsymbol{s_e}$ are the mean height and existence score vectors generated by MHG and ESG, respectively.
Note that the existence score condition is occasionally used alone to show the effect of each condition in the following experiments.
The steps of MHAS are summarized in Algorithm \ref{alg:MHAS}.

\begin{algorithm}
   \caption{Mean Height Aided Suppression (MHAS).}
   \label{alg:MHAS}
\begin{algorithmic}
   \STATE {\bfseries Input:} data $y, h, \boldsymbol{s_e}, \boldsymbol{\Bar{h}}$, hyper-parameters $b_h, \alpha_h, \alpha_s, r$
   \STATE Initialize $m_h=\lceil(y+h)/b_h\rceil$.
   \FOR{$m=m_h-r$ {\bfseries to} $m_h+r$}
   \IF{$|h/\boldsymbol{\bar{h}}[m]-1|\leq \alpha_h$ \and $\boldsymbol{s_e}[m]\geq \alpha_s$}
   \STATE $Break$
   \STATE \textbf{Output}: $y, h$
   \ENDIF
   \ENDFOR
   \STATE \textbf{Output}: None
\end{algorithmic}
\end{algorithm}

\subsection{Choice of Hyper-Parameters}
\label{sec:mhas-choice of parameters}
For higher detection accuracy, pedestrian detectors, such as \cite{liu2019CSP}, commonly employ feature maps downsampled to 1/4 concerning the input image.
This high resolution facilitates the recognition of even small pedestrians.
$\boldsymbol{s_e}$ and $\boldsymbol{\bar{h}}$ keep the same resolution, while $b_h$ is set to 4 to align with the requirements of the pedestrian detectors.
The remaining parameters, namely $\alpha_h$, $\alpha_s$ and $r$ are selected on a case-by-case basis, considering specific scenarios and conditions.

It is observed that the optimal selection of these parameters varies when the proposed method is applied to different detectors or datasets.
This is because the proposed method relies on the detection results, yet different detectors can produce false positives with significantly different patterns.
To achieve robust performance, we implement a case-specific strategy with which appropriate hyper-parameters can be chosen for any detector and dataset.

This strategy evaluates pedestrian detection performance on the pseudo-validation set for each combination of $(\alpha_h, \alpha_s, r)$.
The combination that exhibits the best performance is subsequently chosen for the test set with the specific dataset and detector.
The best performance relies on minimizing the log-average miss rate (MR$^{-2}$) while simultaneously maintaining the highest recall achievable. 
This is because the MR$^{-2}$ can be reduced significantly, however, at the expense of mistakenly rejected true positives.
To ensure that the post-processing phase does not result in additional missed pedestrians or minimizes their occurrence, we prioritize achieving a high recall as a prerequisite criterion. 
Additionally, during the evaluation, we aim for a low value of (MR$^{-2}$) to strike a balance between reducing missed detections and maintaining overall robust performance.

As the two most commonly used pedestrian datasets (Caltech and Citypersons) do not contain a validation set or realise the test set online, we create the pseudo-validation set as a substitution for the real validation set.
This is done by randomly choosing images from the training set.
For the Caltech dataset, we randomly choose video sequences from 6 training sets (Set00: seq 5, 4, 8; Set01: seq 0, 2, 4, 5; Set02: seq 3, 11, 5, 6, 7; Set03: seq 11, 2, 4, 6; Set04: seq 10, 2, 5, 9; Set05: seq 0, 7, 3, 1) as the pseudo-validation set. 
For the Citypersons dataset, we randomly choose 500 training images as the pseudo-validation set.

\section{Experiments}
\label{sec: exp results}
\subsection{Settings}
\label{sec: experiment settings}
\subsubsection{Datasets}
Caltech Pedestrian Dataset \cite{dollar2011pedestrian} contains approximately 250k training images with a resolution of 480x640 from a 10-hour video taken in a single urban city.
In our experiments, we use the augmented dataset which contains 42782 training images and 4024 test images with new annotations \cite{zhang2016far} in our experiments.
Citypersons Dataset \cite{zhang2017citypersons} contains 2975 training images with a resolution of 1024x2048 taken from 18 cities, three seasons and different weather conditions.
The validation set contains 500 images taken across three cities.
A Pseudo-Validation Dataset is created to find optimal parameters in MHAS.
It is established by randomly selecting part of the training images.
For the Caltech dataset, we randomly choose 24 video sequences from 6 training sets (Set00: seq 5, 4, 8; Set01: seq 0, 2, 4, 5; Set02: seq 3, 11, 5, 6, 7; Set03: seq 11, 2, 4, 6; Set04: seq 10, 2, 5, 9; Set05: seq 0, 7, 3, 1) as the pseudo-validation set. Overall, 14793 frames are selected.
For the Citypersons dataset, we randomly choose 500 training images as the pseudo-validation set.

\subsubsection{Evaluation settings} 
The pedestrians are usually divided into four subsets (Table \ref{table: experiment settings}) according to their height and visibility.
Visibility is the ratio of the visible box area to the complete box area.
The performance of the pedestrian detector is evaluated by the log-average Miss Rate (denoted as MR$^{-2}$) over the False Positive Per Image (FFPI) range of [10$^{-2}$, 10$^0$].
A lower MR$^{-2}$ value signifies superior performance.
In addition, the recall rate, as well as the number of true and false positives, are occasionally employed to provide a more detailed performance analysis.

\begin{table}[!t]
\caption{Experiment settings of four subsets of pedestrian datasets. The height is counted in pixels. The visibility of a person is determined by the proportion of its visible area to its complete size.}
\centering
\begin{tabular}{|c||c|c|}
\hline
Subset     & Height      & Visibility     \\ 
\hline
Reasonable & {[}50, inf] & {[}0.65, inf]  \\
Small      & {[}50, 75]  & {[}0.65, inf]  \\
Heavy      & {[}50, inf] & {[}0.2, 0.65]  \\
All        & {[}20, inf] & {[}0.2, inf]   \\
\hline
\end{tabular}
\label{table: experiment settings}
\end{table}

\subsection{Ablation Study}
\label{sec: ablation}
\subsubsection{Performance of MHAS}
To evaluate the effectiveness of the proposed MHAS method, we compared it with the baseline results on two datasets as shown in Table \ref{table: ablation-caltech-csp+vit} and \ref{table: ablation-cp-csp+vit}. 
The baseline results in this paper are produced by the CSP detector combined with a deformable encoder neck \cite{csp+vit} unless otherwise specified.
The existence scores and mean heights are pre-generated with the ESG and the manual approach, respectively.
The performance metrics of the MHAS method, using only existence scores and in conjunction with mean heights, are presented.
Results demonstrate that the proposed method improves the detection across all subsets in both datasets.
It's noteworthy that this improvement is largely attributed to the combined utilization of both existence scores and mean heights.
\begin{table*}[!t]
\caption{
Ablation study on the performance of the proposed MHAS algorithm on the Caltech dataset.
The algorithm is examined utilizing only the existence score principle and combined with additional manual mean heights.
The third and fourth columns are the results of the pseudo-validation set, and the rest are the test set.
The best results on the test set are in \textbf{bold}.}
\label{table: ablation-caltech-csp+vit}
\centering
\begin{tabular}{|c|c||c|c||c|c|c|}
\hline
Method   & $\alpha_{h}, \alpha_{s}, r$ & Reasonable & All    & Reasonable & Heavy  & All    \\ \hline
Baseline & \textbackslash{}            & 9.628      & 17.322 & 4.141      & 44.688 & 57.883 \\
+Existence & \textbackslash{}, 0.04, 1   & 5.875 & 10.350 & 3.922(0.219$\downarrow$) & 43.678(1.01$\downarrow$)  & 57.311(0.572$\downarrow$) \\
++Mean H   & 0.8, 0.04, 1 & 6.105 & 10.542 & \textbf{3.893}(\textbf{0.248}$\downarrow$) & \textbf{43.241}(\textbf{1.447}$\downarrow$) & \textbf{57.158}(\textbf{0.725}$\downarrow$) \\ \hline
\end{tabular}%
\end{table*}

\begin{table*}[!t]
\caption{Ablation study on the performance of the proposed MHAS algorithm on Citypersons dataset.
The algorithm is examined utilizing only the existence score principle and combined with additional manual mean heights.
The third and fourth columns are the results of the pseudo-validation set, and the rest are the test set.
The best results on the validation set are in \textbf{bold}.}
\label{table: ablation-cp-csp+vit}
\centering
\begin{tabular}{|c|c||c|c||c|c|c|c|}
\hline
Method   & $\alpha_{h}, \alpha_{s}, r$ & Reasonable & Small & Reasonable & Small & Heavy & All   \\ \hline
Baseline & \textbackslash{}            & 35.18      & 25.17 & 10.63      & 14.88 & 36.67 & 36.39 \\
+Existence &
  \textbackslash{}, 0.05, 5 &
  35.12 &
  25.00 &
  10.42(0.21$\downarrow$) &
  \textbf{14.31(0.57$\downarrow$)} &
  35.82(0.85$\downarrow$) &
  35.99(0.4$\downarrow$) \\
++Mean H &
  0.5, 0.05, 16 &
  35.19 &
  25.00 &
  \textbf{10.28(0.35$\downarrow$)} &
  \textbf{14.31(0.57$\downarrow$)} &
  \textbf{35.80(0.87$\downarrow$)} &
  \textbf{35.93(0.46$\downarrow$)} \\ \hline
\end{tabular}%
\end{table*}

\subsubsection{Effectiveness of case-specific choice}
Table \ref{table: ablation-caltech-optim-dataset} compares the performance with the parameters in MHAS optimized on the test set or the pseudo-validation set.
Overfitting parameters on the test set aims to achieve the theoretical best performance of the proposed method.
The actual performance is achieved by optimizing MHAS on the pseudo-validation set.
Notably, the actual performance closely aligns with the theoretical performance, indicating the effectiveness and robustness of the proposed method.

\begin{table}[htbp]
\centering
\caption{
Study of the effectiveness of the case-specific choice.
The performance with hyper-parameters are optimized on the test set or pseudo-validation set (Pseudo Val.). 
The notation 'E' and 'MH' refer to using existence score and mean height principles, respectively.
The notation 'D' refers to the dataset on which hyper-parameters are optimized. 
The performance is evaluated on the Caltech test set.}
\begin{tabular}{|c|c|c||c|c|c|}
\hline
E      & MH     & D           & Reasonable & Heavy & All   \\ \hline
\xmark & \xmark & \textbackslash{}           & 4.14       & 44.69 & 57.88 \\
\cmark & \xmark & Test        & 3.92       & 43.68 & 57.31 \\
\cmark & \xmark & Pseudo Val. & 3.92       & 43.68 & 57.31 \\
\cmark & \cmark & Test        & 3.87       & 43.02 & 57.08 \\
\cmark & \cmark & Pseudo Val. & 3.89       & 43.24 & 57.16 \\ \hline
\end{tabular}%
\label{table: ablation-caltech-optim-dataset}
\end{table}

\subsubsection{Selection of pseudo-validation set}
To investigate the impact of image selection on the choice of hyperparameters and the performance of MHAS, we generate a second pseudo-validation set based on the Caltech and Citypersons training sets, respectively.
For the Caltech dataset, the second pseudo-validation set is made up of 24 video sequences from 6 training sets (Set00: seq 2, 6, 12; Set01: seq 1, 2, 3, 5; Set02: seq 0, 4, 3, 9, 10; Set03: seq 0, 4, 8, 12; Set04: seq 0, 7, 9, 11; Set05: seq 0, 5, 9, 11) totalling 14503 frames.
For the Citypersons dataset, another 500 training images are randomly chosen to form the second pseudo-validation set.
Table \ref{table: different pseudo val sets-cal} compares the detection results on Caltech while Tables \ref{table: different pseudo val sets-cp} and \ref{table: different pseudo val sets-cp-tpfp} present the results measured in terms of MR$^{-2}$ and the number of true (false) positives respectively.
The MR$^{-2}$ results of the baseline CSP detector are similar in two pseudo sets.
This indicates that the selected images sufficiently represent the entire training set.
While the whole training set could be utilized for determining the most optimal hyperparameters, a small unbiased sample is employed to save computational time.
Results on both datasets demonstrate that different pseudo-validation sets yield varying degrees of stable improvement.
Specifically, the second pseudo-validation set leads to a more noticeable decrease in MR$^{-2}$ on each subset of the Caltech dataset compared to the first set.
Conversely, for the Citypersons dataset, both sets produce very close detection results.

\begin{table*}[!t]
\caption{Results of MHAS aided detection with coefficients optimized on different pseudo validation sets.
The performance is assessed on the Caltech dataset in both the standalone separate existence score component and in combination with the manual mean height component.
The third and fourth columns are the results of the pseudo-validation set, and the rest are the test set.
For the pseudo-validation set, the MR$^{-2}$ (left) and recall (right) are presented.
'P. Val' stands for pseudo-validation set.}
\label{table: different pseudo val sets-cal}
\centering
\begin{tabular}{|c|c||c|c|c|c|c|}
\hline
Detector    & P. Val         & Reasonable          & All                 & Reasonable & Heavy  & All    \\ \hline
CSP         & \multirow{3}{*}{1} & 2.471$\vert$ 99.812 & 3.952$\vert$ 99.744 & 4.816      & 47.380 & 59.310 \\
+Existence  &                    & 2.471$\vert$ 99.812 & 3.548$\vert$ 99.693 & 4.303      & 46.913 & 58.414 \\
++Mean H &                    & 2.295$\vert$ 99.812 & 3.623$\vert$ 99.693 & 4.259      & 46.913 & 58.381 \\ \hline
CSP         & \multirow{3}{*}{2} & 3.31$\vert$ 99.567  & 4.06$\vert$ 99.613  & 4.816      & 47.380 & 59.310 \\
+Existence  &                    & 3.20$\vert$ 99.533  & 3.909$\vert$ 99.582 & 4.172      & 46.456 & 58.508 \\
++Mean H &                    & 3.229$\vert$ 99.550 & 3.961$\vert$ 99.598 & 3.982      & 44.056 & 57.462 \\ \hline
\end{tabular}%
\end{table*} 
\begin{table*}[!t]
\caption{Results of MHAS aided detection with coefficients optimized on different pseudo validation sets.
The performance is assessed on the Citypersons dataset in both the standalone separate existence score component and in combination with the manual mean height component.
The third and fourth columns are the results of the pseudo-validation set, and the rest are the validation set.
For the pseudo-validation set, the MR$^{-2}$ (left) and recall (right) are presented.
'P. Val' stands for pseudo-validation set.}
\label{table: different pseudo val sets-cp}
\centering
\begin{tabular}{|c|c||c|c|c|c|c|c|}
\hline
Detector   & P. Val         & Reasonable           & Small                & Reasonable & Small  & Heavy  & All    \\ \hline
CSP        & \multirow{3}{*}{1} & 18.73$\vert$ 91.41   & 17.3$\vert$ 93.97    & 10.904     & 14.77  & 41.35  & 37.56  \\
+Existence &                    & 18.697$\vert$ 91.363 & 17.143$\vert$ 93.968 & 10.735     & 14.297 & 40.958 & 37.206 \\
++Mean H   &                    & 18.687$\vert$ 91.363 & 17.143$\vert$ 93.968 & 10.735     & 14.297 & 40.958 & 37.206 \\ \hline
CSP        & \multirow{3}{*}{2} & 17.749$\vert$ 90.00  & 19.37$\vert$ 92.00   & 10.904     & 14.77  & 41.35  & 37.56  \\
+Existence &                    & 17.717$\vert$ 89.90  & 19.373$\vert$ 92.00  & 10.735     & 14.297 & 40.807 & 37.164 \\
++Mean H   &                    & 17.717$\vert$ 89.90  & 19.373$\vert$ 92.00  & 10.735     & 14.297 & 40.807 & 37.164 \\ \hline
\end{tabular}%
\end{table*}
\begin{table*}[!t]
\caption{Comparison of the number of true and false positives of MHAS aided detection with coefficients optimized on different pseudo validation sets.
The performance is assessed on the Citypersons dataset in both the standalone separate existence score component and in combination with the manual mean height component.
The third and fourth columns are the results obtained on the pseudo-validation set, and the rest are the results on the validation set.
For each subset, the number of true positives (left) and false positives (right) are presented.
'P. Val' stands for pseudo-validation set.}
\label{table: different pseudo val sets-cp-tpfp}
\centering
\begin{tabular}{|c|c||c|c|c|c|c|c|}
\hline
Detector   & P. Val & Reasonable       & Small          & Reasonable       & Small           & Heavy           & All              \\ \hline
CSP & \multirow{3}{*}{1} & 2040$\vert$ 1643 & 601$\vert$ 879  & 1528$\vert$ 1787 & 340$\vert$ 1142 & 635$\vert$ 1413 & 2385$\vert$ 2135 \\
+Existence &        & 2037$\vert$ 1594 & 601$\vert$ 853 & 1528$\vert$ 1738 & 340$\vert$ 1105 & 634$\vert$ 1364 & 2383$\vert$ 2083 \\
++Mean H   &        & 2037$\vert$ 1588 & 601$\vert$ 853 & 1528$\vert$ 1735 & 340$\vert$ 1105 & 634$\vert$ 1361 & 2383$\vert$ 2080 \\ \hline
CSP & \multirow{3}{*}{2} & 1950$\vert$ 1651 & 607$\vert$ 1015 & 1528$\vert$ 1787 & 340$\vert$ 1142 & 635$\vert$ 1413 & 2385$\vert$ 2135 \\
+Existence &        & 1948$\vert$ 1620 & 607$\vert$ 998 & 1528$\vert$ 1741 & 340$\vert$ 1107 & 634$\vert$ 1368 & 2383$\vert$ 2087 \\
++Mean H   &        & 1948$\vert$ 1618 & 607$\vert$ 998 & 1528$\vert$ 1737 & 340$\vert$ 1107 & 634$\vert$ 1364 & 2383$\vert$ 2083 \\ \hline
\end{tabular}%
\end{table*}
\subsubsection{Importance of existence scores}
The existence scores are required in conjunction with mean heights for a complete MHAS module.
This inference stems from the observation that performance can be significantly enhanced with the application of existence scores, in addition to the exclusive use of mean heights, as demonstrated in Table \ref{table: ablation-importance of existence scores}.
This arises because the generated mean heights are non-zero at any level under the horizon line.
In this case, false positives that coincidentally fulfil the mean height condition are mistakenly retained.
To overcome this problem, the existence score is produced to predict the probability of a level containing pedestrians.
Scores that fall below a pre-set threshold suggest the absence of pedestrians at corresponding levels, leading to the rejection of erroneous predictions. 
This condition complements the mean height criterion in rejecting bounding boxes above the horizon line, serving as a dual verification system.
Table \ref{table: ablation-importance of existence scores} shows that with the inclusion existence scores, the MR$^{-2}$ is decreased significantly even though the mean height condition has already improved the performance.
Thus, the mean height condition is recommended to be utilized together with the existence score condition to achieve optimal performance. 

\begin{table}[!t]
\caption{The importance of the assistance of the existence score compared with the sole use of manual mean heights.}
\label{table: ablation-importance of existence scores}
\centering
\begin{tabular}{|c||c|c|c|}
\hline
Method      & Reasonable & Heavy  & All    \\ \hline
Baseline    & 4.14      & 44.67 & 57.88 \\
+Mean H     & 4.02      & 44.10 & 57.47 \\
++Existence & 3.87       & 43.02 & 57.08    \\ \hline
\end{tabular}%
\end{table}
\subsubsection{MHG approaches}
As mentioned in Section \ref{sec: mhg}, manual and perspective prediction methods can be employed to generate the mean heights.
The manual approach is suitable for images with consistent eye-level perspectives, while the perspective prediction method is suitable for images with changing eye levels.
Table \ref{table: predmh-cal-csp+vit} and \ref{table: predmh-cp-csp+vit} present the results of MHAS with these two approaches on the Caltech test set and Citypersons validation set, respectively.
Both approaches yield closely aligned and improved performance.
This is because the Caltech and Citypersons datasets consist of video sequences captured from fixed camera positions on specific vehicles, resulting in relatively stable eye levels.
On these two datasets, the manual method slightly outperforms the perspective prediction method when combined with the baseline detector.
A possible reason is that there might not be a sufficient number of qualified images available to train an accurate perspective coefficient predictor.
Only 2348 images are selected from the Caltech training set and 1143 from the Citypersons training set with a mean relative error of 0.4.
To investigate the influence of the training set on the perspective predictor, Table \ref{table: predmh data maxErr-cp-csp+vit} presents the results with different mean relative errors on the Citypersons validation set.
The smaller the error, the fewer qualified images for training and relatively more accurate ground truth perspective coefficients.
The larger the error, the more flawed samples are included.
For example, the numbers of qualified Citypersons images are 1335, 1143 and 827, with mean relative errors of infinity, 0.4 and 0.1.
Interestingly, both large and small errors appear to result in relatively higher MR$^{-2}$ particularly for the small and heavy subsets.
This is because small errors lead to a lack of training images, making the predictor prone to overfitting, while large errors introduce more ambiguous information due to inaccurate ground truths.
\begin{table}[!t]
\caption{Comparison of the two mean height generation approaches embedded in the MHAS algorithm.
The mean heights are generated through manual computation or linear perspective.
Results are obtained on the Caltech test set.}
\label{table: predmh-cal-csp+vit}
\centering
\begin{tabular}{|c||c|c|c|}
\hline
Method   & Reasonable & Heavy  & All    \\ \hline
Baseline & 4.141      & 44.688 & 57.883 \\
Manual   & 3.893      & 43.241 & 57.158 \\
Pred     & 3.903      & 43.570 & 57.232 \\ \hline
\end{tabular}%
\end{table}
\begin{table}[!t]
\caption{Comparison of the two mean height generation approaches embedded in the MHAS algorithm.
The mean heights are generated through manual computation or linear perspective.
Results are obtained on the Citypersons validation set.}
\label{table: predmh-cp-csp+vit}
\centering
\begin{tabular}{|c||c|c|c|c|}
\hline
Method   & Reasonable & Small & Heavy & All   \\ \hline
Baseline & 10.63      & 14.88 & 36.67 & 36.39 \\
Manual   & 10.28      & 14.31 & 35.80 & 35.93 \\
Pred     & 10.35     & 14.21 & 35.86 & 36.15 \\ \hline
\end{tabular}%
\end{table}
\begin{table}[!t]
\caption{Influence of the maximum relative error used in the perspective coefficient approach.}
\label{table: predmh data maxErr-cp-csp+vit}
\centering
\begin{tabular}{|c||c|c|c|c|}
\hline
Max err & Reasonable & Small & Heavy & All   \\ \hline
0.2     & 10.35      & 14.31 & 35.92 & 35.99 \\
0.4     & 10.35      & 14.21 & 35.86 & 36.15 \\
inf     & 10.35      & 14.28 & 35.90 & 36.00 \\ \hline
\end{tabular}%
\end{table}

\subsubsection{Robustness}
\begin{figure*}[!t]
\centering
\subfloat[]{\includegraphics[width=1.7in]{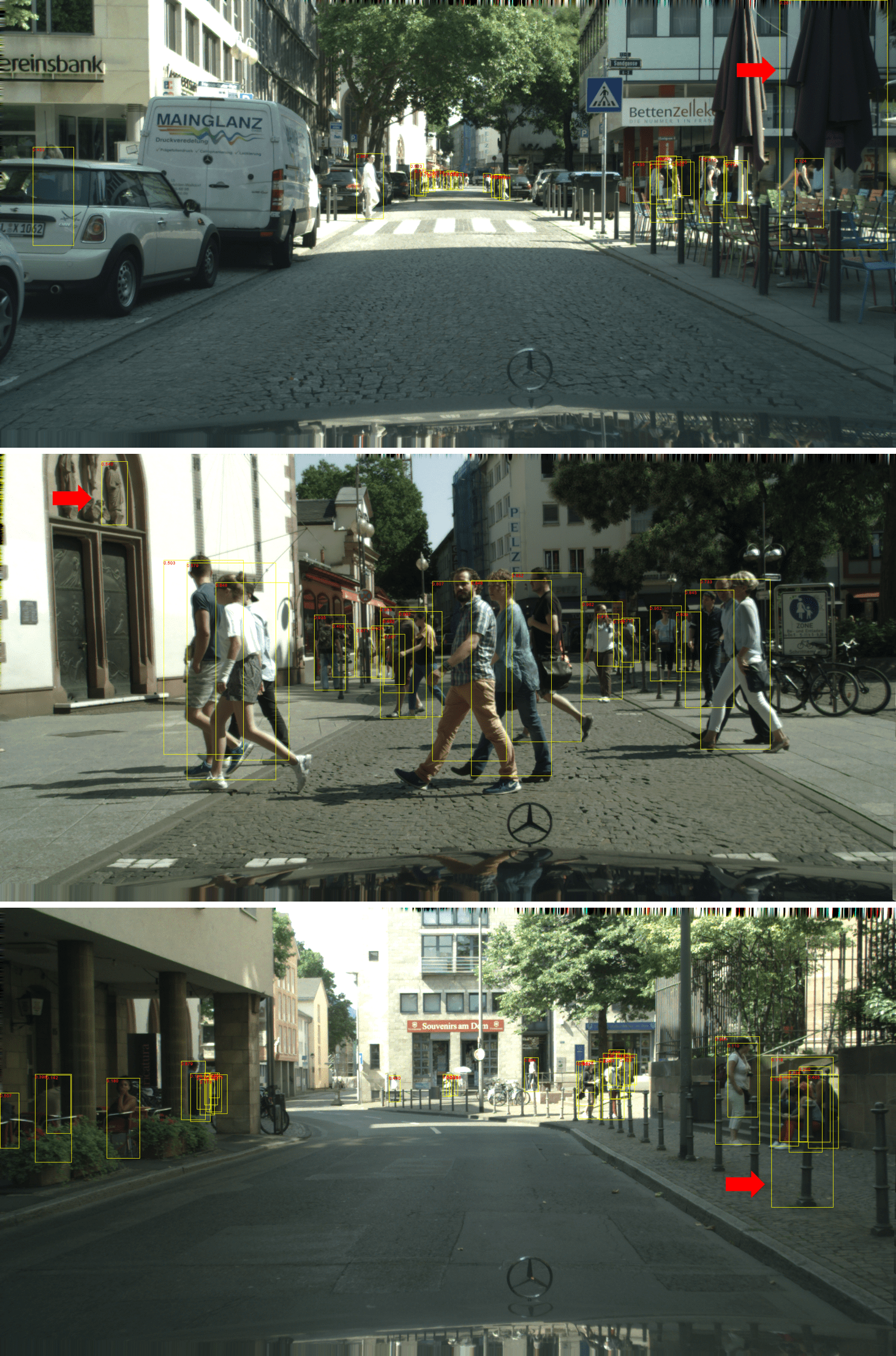}}
\hspace{-0.cm}
\subfloat[]{\includegraphics[width=1.7in]{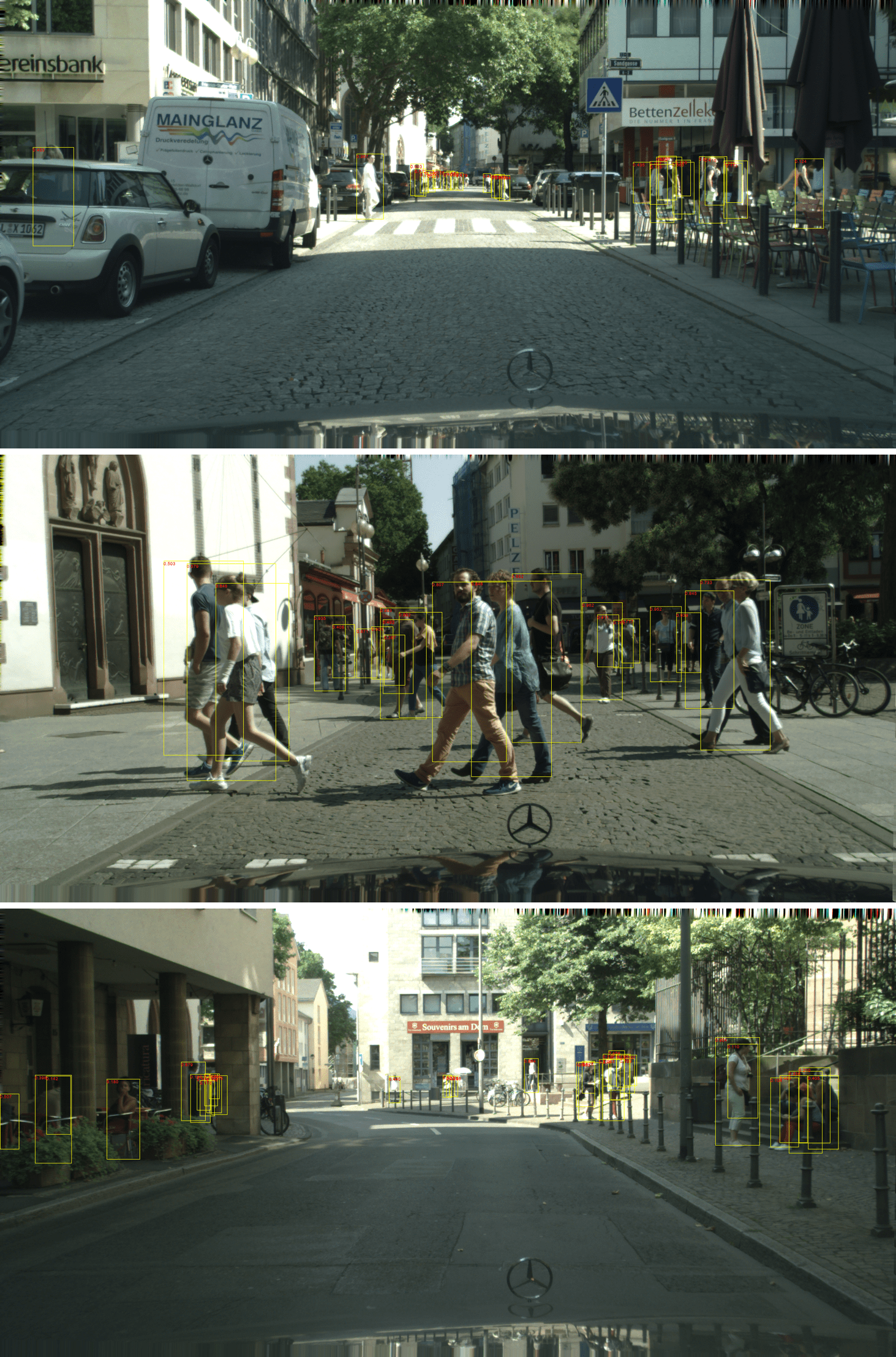}}
\hspace{0.2cm}
\subfloat[]{\includegraphics[width=1.7in]{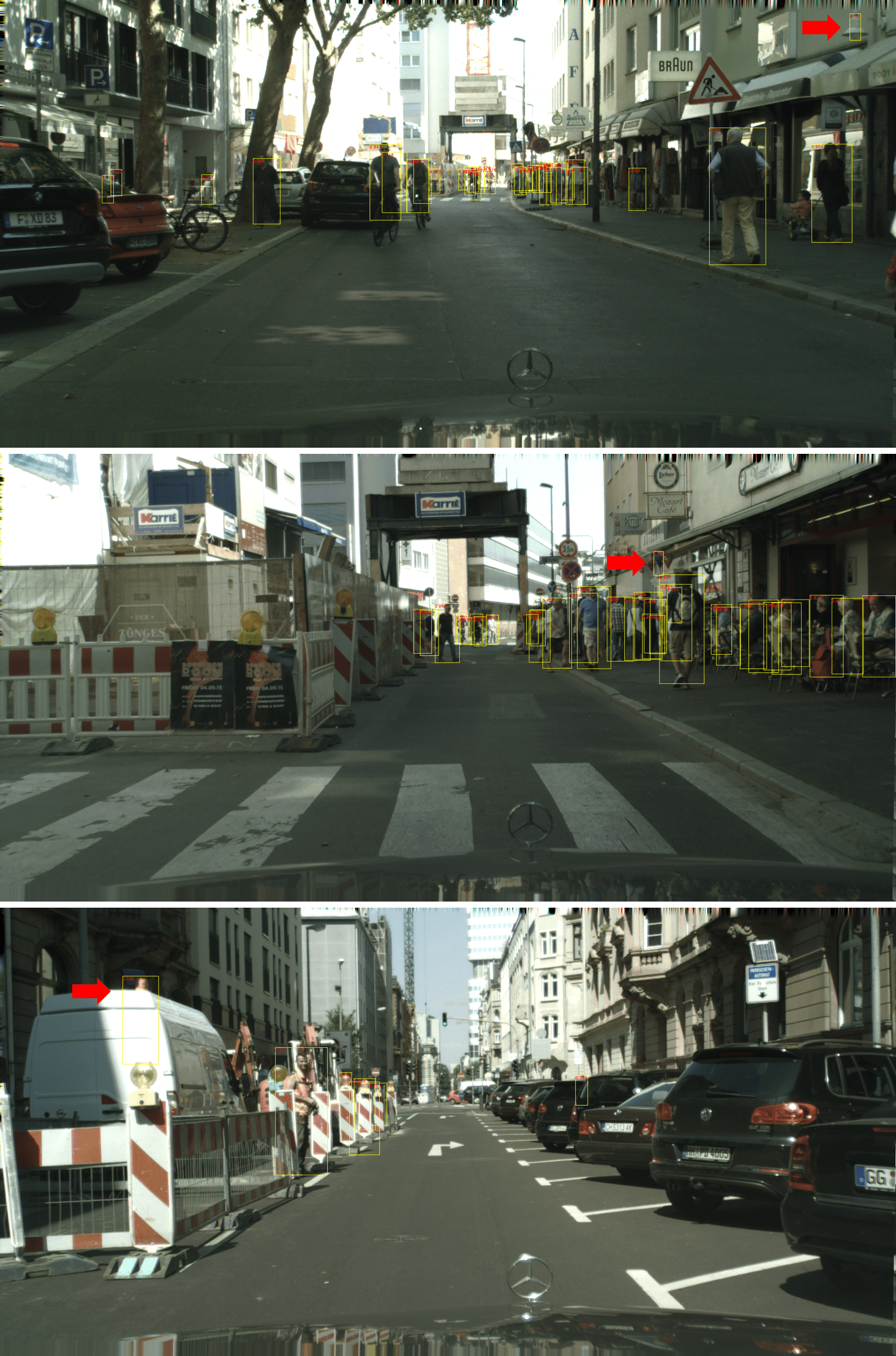}}
\hspace{-0.cm}
\subfloat[]{\includegraphics[width=1.7in]{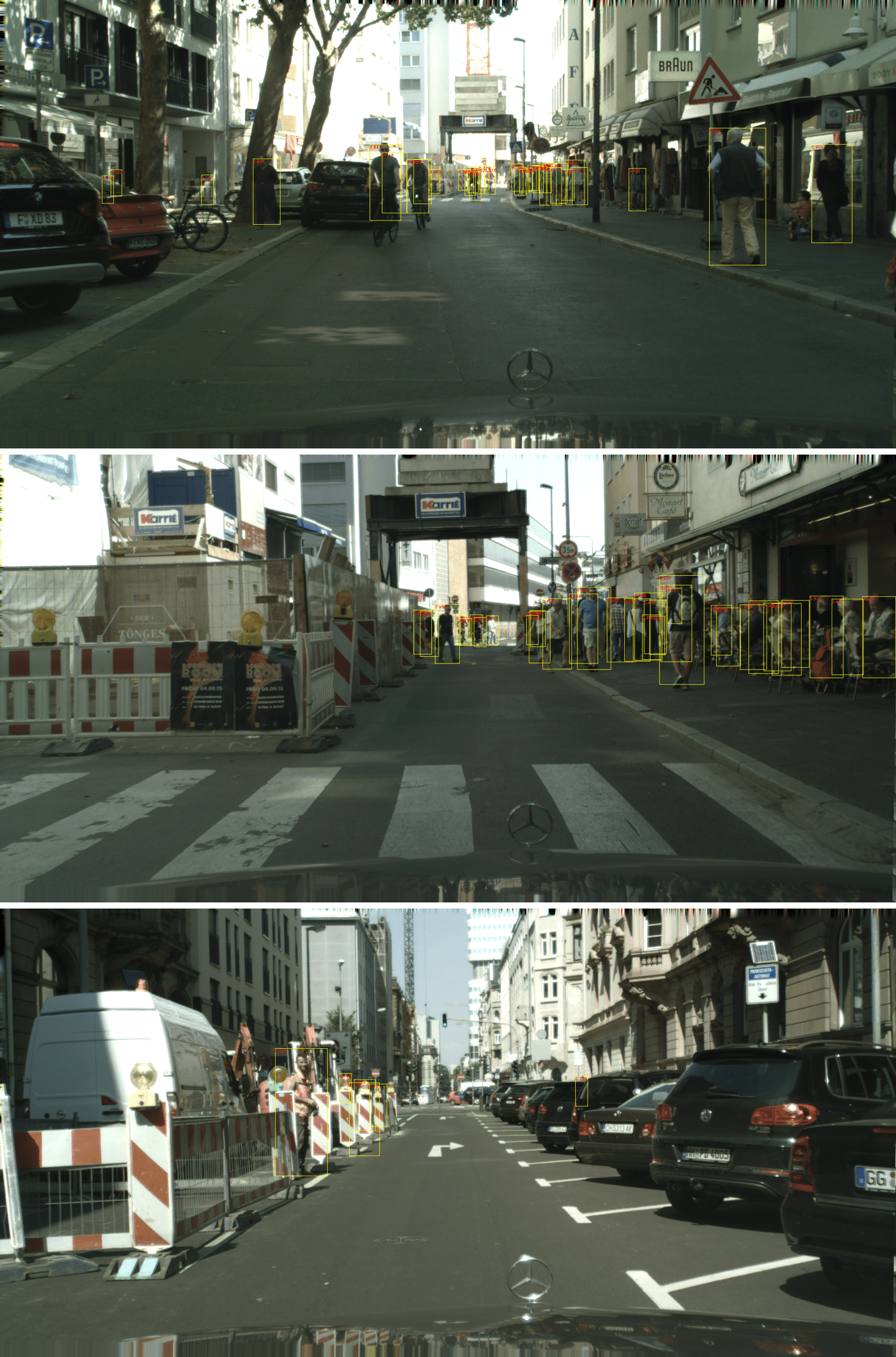}}
\caption{Detection results of CSP with or without MHAS. (a), (c) Pedestrians detected by CSP. Arrows mark the false positive bounding boxes which are rejected after MHAS. (b), (d) Detection results of CSP with the aid of MHAS correspond to (a) and (c), respectively.}
\label{fig:MHAS-det-results-csp+mhas}
\end{figure*}

\begin{figure*}[!t]
\centering
\subfloat[]{\includegraphics[width=1.7in]{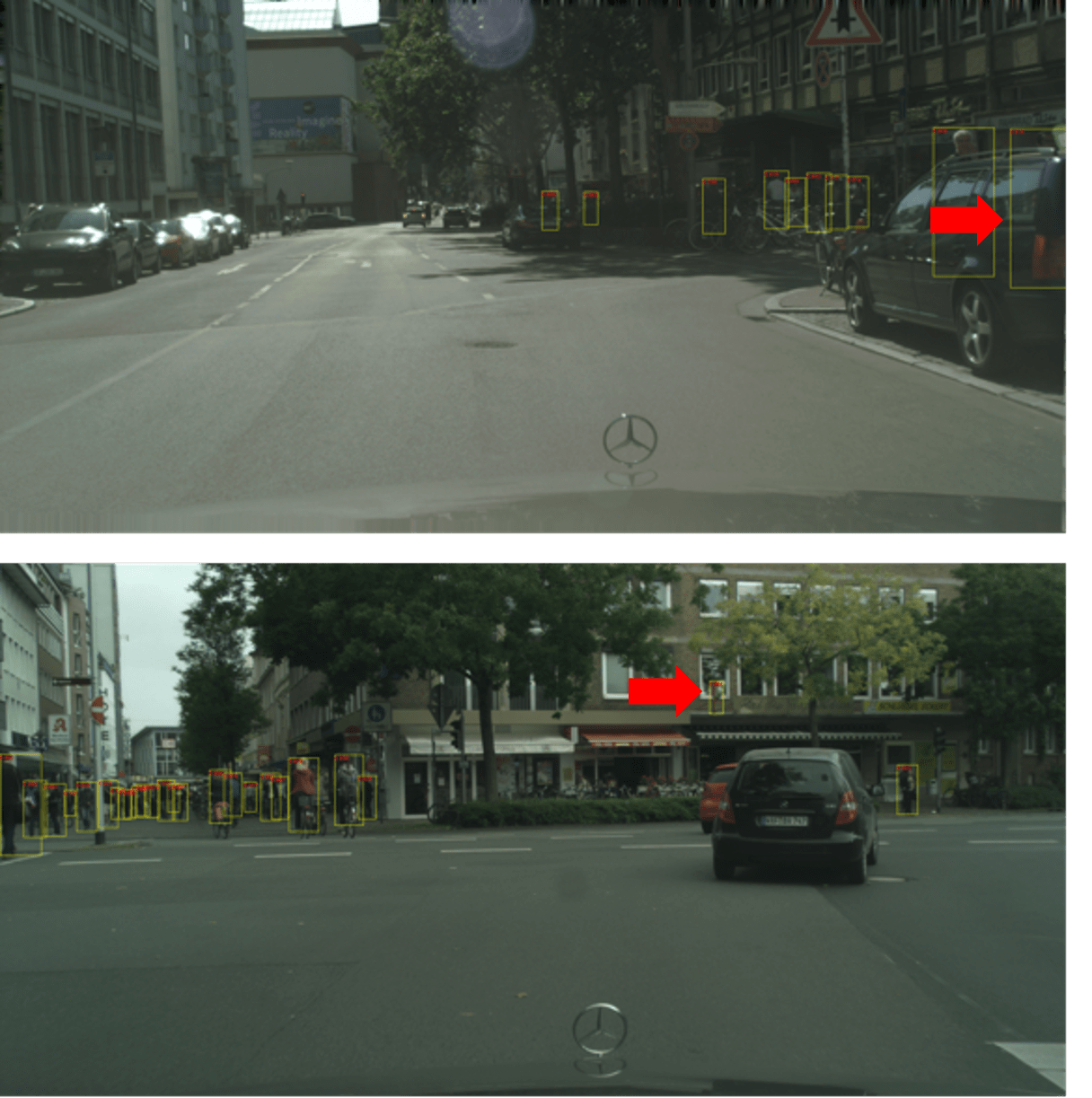}}
\hspace{-0.2cm}
\subfloat[]{\includegraphics[width=1.7in]{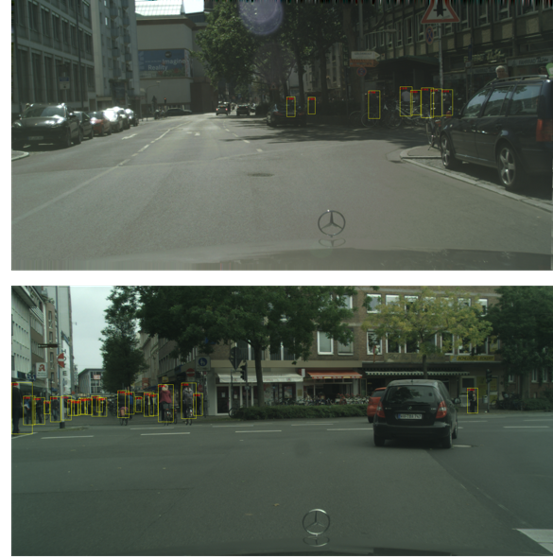}}
\hspace{0.2cm}
\subfloat[]{\includegraphics[width=1.7in]{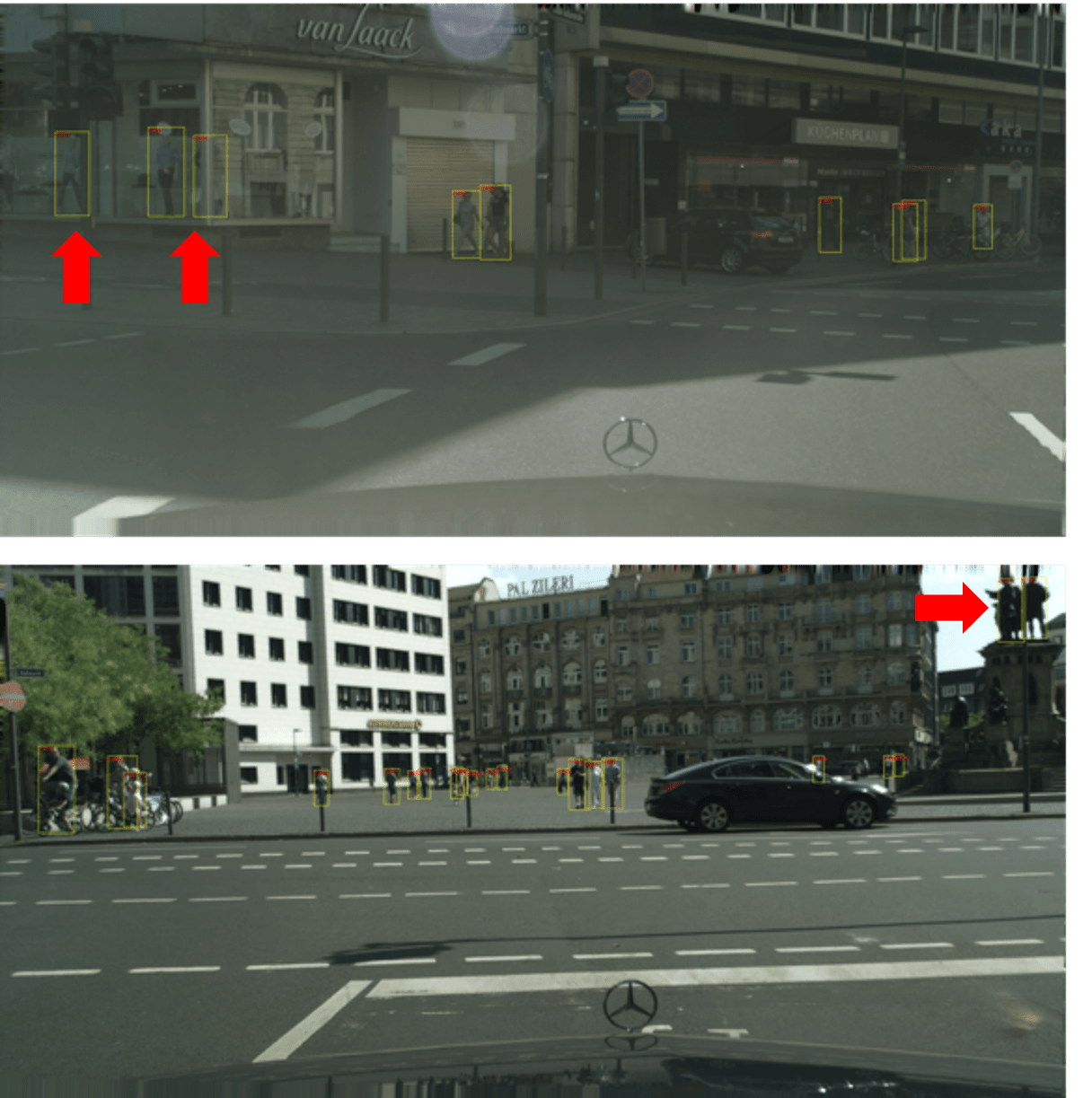}}
\hspace{-0.2cm}
\subfloat[]{\includegraphics[width=1.7in]{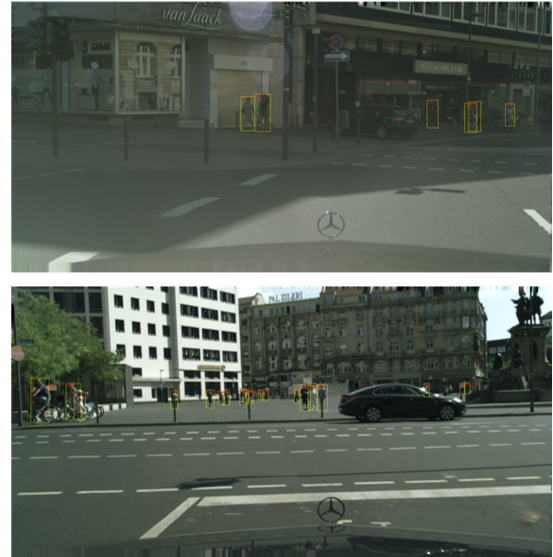}}
\caption{Detection results of APD with or without MHAS. (a), (c) Pedestrians detected by APD. Arrows mark the false positive bounding boxes which are rejected after MHAS. (b), (d) Detection results of APD with the aid of MHAS correspond to (a) and (c), respectively.}
\label{fig:MHAS-det-results-apd+mhas}
\end{figure*}

The robustness of the complete MHAS method is evaluated when it is applied to various state-of-the-art pedestrian detectors on the Caltech test set (Table \ref{table: ablation-robustenss-caltech-detectors}) and the challenging Citypersons validation set (Table \ref{table: ablation-robustness-cp-detectors}).
For the first dataset, the single-stage pedestrian detector CSP \cite{liu2019CSP} and two-stage Cascade R-CNN \cite{cascade-rcnn} are presented.
CSP is particularly designed for pedestrian detection. 
It makes a one-round prediction, so the results usually contain more false positives.
The Cascade R-CNN utilizes multiple rounds of region-wise feature extraction and bounding box refinement.
It serves as a powerful benchmark by producing more accurate predictions with reduced false positives.
Results show that the proposed MHAS significantly improves the performance of single-stage detectors, namely CSP \cite{liu2019CSP} and APD \cite{APD} across all subsets on both Caltech and Citypersons datasets.
For Citypersons, results obtained on APD \cite{APD} and MGAN \cite{mgan2019} are presented alongside CSP.
APD uses a loss-based attribute-aware NMS during the post-processing stage for pedestrian detection.
MGAN is a visible mask-guided two-stage pedestrian detector.
Fig. \ref{fig:MHAS-det-results-csp+mhas} and \ref{fig:MHAS-det-results-apd+mhas} present some sample detection results of the CSP and APD detectors and their combination with the MHAS algorithm.

For single-stage detectors, like CSP and APD, the performance is improved significantly with MHAS on both datasets.
Although two-stage detectors, like Cascade R-CNN and MGAN, have achieved accurate detection by taking advantage of the region proposal network and prediction fine-tuning, the proposed method still achieves improvements on specific subsets.
It is worth mentioning that the improvement on two-stage detectors is observed to be smaller than on single-stage ones.
This discrepancy can be attributed to the fact that the former detectors have effectively suppressed a majority of false positives, leaving only a few for MHAS to address.
They also implicitly learn better the common size of bounding boxes at different levels compared to single-stage detectors.
This indicates that part of the mean height condition filtering has been done within the detector itself.
Consequently, the reduction of false positives yields limited benefits for two-stage detectors, whereas an increase in true positives proves more advantageous.
Therefore, reducing false positives has limited benefits; and it benefits more from increasing the number of true positives.

\begin{table*}[!t]
\caption{Robustness evaluation with the proposed MHAS algorithm implemented on various pedestrian detectors on the Caltech dataset. 
The second and third columns are the results of the pseudo-validation set, and the rest are the results of the test set.
For the pseudo-validation set, the MR$^{-2}$ (left) and recall (right) are presented.
Improved MR$^{-2}$ on test set is in \textbf{bold}.}
\label{table: ablation-robustenss-caltech-detectors}
\centering
\begin{tabular}{|c||c|c||c|c|c|}
\hline
Method        & Reasonable          & All                  & Reasonable     & Heavy  & All    \\ \hline
CSP           & 2.471$\vert$ 99.812 & 3.952$\vert$ 99.744  & 4.816          & 47.380 & 59.310 \\
+MHAS & 2.295$\vert$ 99.812 & 3.623$\vert$ 99.693 & \textbf{4.259} & \textbf{46.913} & \textbf{58.381} \\ \hline
Cascade R-CNN & 8.204$\vert$ 97.217 & 12.107$\vert$ 97.210 & 1.477          & 22.108 & 25.482 \\
+MHAS         & 8.371$\vert$ 97.061 & 11.940$\vert$ 97.082 & \textbf{1.439} & 23.118 & 29.079 \\ \hline
\end{tabular}%
\end{table*}

\begin{table*}[!t]
\caption{Robustness evaluation with the proposed MHAS algorithm implemented on various pedestrian detectors on the Citypersons dataset. 
The second and third columns are the results of the pseudo-validation set, and the rest are the 
 results of the validation set.
For the pseudo-validation set, the MR$^{-2}$ (left) and recall (right) are presented.
Improved MR$^{-2}$ on the validation set is in \textbf{bold}.}
\label{table: ablation-robustness-cp-detectors}
\centering
\begin{tabular}{|c||c|c||c|c|c|c|}
\hline
Method & Reasonable          & Small               & Reasonable     & Small          & Heavy          & All            \\ \hline
CSP    & 18.73$\vert$ 91.41 & 17.31$\vert$ 93.97  & 10.90          & 14.77          & 41.35          & 37.56          \\
+MHAS  & 18.73$\vert$ 91.32 & 17.14$\vert$ 93.97  & \textbf{10.74} & \textbf{14.30} & \textbf{40.78} & \textbf{37.16} \\ \hline
APD    & 14.57$\vert$ 93.16  & 15.08$\vert$ 91.59  & 9.70           & 16.62          & 41.34          & 36.42          \\
+MHAS  & 14.78$\vert$ 92.98  & 15.08$\vert$ 91.59  & \textbf{9.41}  & \textbf{15.99} & \textbf{40.99} & \textbf{35.86} \\ \hline
MGAN   & 26.49$\vert$ 80.37  & 26.56$\vert$ 79.37  & 11.29          & 15.04          & 52.58          & 42.51          \\
+MHAS  & 26.53$\vert$ 80.32  & 26.56$\vert$ 79.37  & \textbf{11.27} & 15.04 & \textbf{52.38} & \textbf{42.48} \\ \hline
\end{tabular}%
\end{table*}


\subsection{Comparison with state-of-the-art}
Tables \ref{table: sota-caltech} and \ref{table: sota-cp-full} compare the performance of the proposed MHAS module with other state-of-the-art pedestrian detectors on the Caltech test set and Citypersons validation set, respectively.
Remarkably, despite being a straightforward post-processing algorithm, MHAS achieves the best performance among the listed detectors on both datasets.
This impressive outcome not only highlights the effectiveness of mean height aided post-processing but also underscores the potential of leveraging the distinctive prior knowledge available in pedestrian datasets.
\begin{table}[!t]
\caption{Comparison with the state-of-the-art pedestrian detectors on Caltech test set.
The best results are in \textbf{bold}.}
\label{table: sota-caltech}
\centering
\begin{tabular}{|c||c|}
\hline
Method                            & Reasonable    \\ \hline
Faster R-CNN \cite{faster-r-cnn}  & 8.7          \\
ALFNet \cite{ALF}                 & 8.1          \\
RPN+BF\cite{RPN+BF}               & 7.3          \\
RepLoss\cite{reploss}             & 5.0          \\
CSP \cite{liu2019CSP}             & 4.5          \\
CSP+Detr \cite{csp+vit}           & 4.1          \\
KGSNet \cite{zhang2020kgsnet}     & 3.9          \\
JointDet \cite{jointDet}          & 3.0          \\
PedHunter \cite{chi2020pedhunter} & 2.3          \\
Cascade R-CNN \cite{cascade-rcnn} & 1.5          \\ \hline
CSP+MHAS (ours)                   & 4.3          \\
CSP+Detr+MHAS (ours)              & 3.9          \\
Cascade R-CNN+MHAS (ours)         & \textbf{1.4} \\ \hline
\end{tabular}%
\end{table}

\begin{table}[!t]
\caption{Comparison with the state-of-the-art pedestrian detectors on Citypersons test set. \textsuperscript{1} Visibility of [0, 0.65] instead of [0.2, 0.65]. 
Results of heavy and all subsets are included.
The best results are in \textbf{bold}.}
\label{table: sota-cp-full}
\centering
\begin{tabular}{|c||c|c|c|}
\hline
Method                                & Reasonable   & \multicolumn{1}{c|}{Heavy} & All           \\ \hline
FRCNN \cite{zhang2017citypersons}     & 15.4         & -                          & -             \\
FRCNN+Seg \cite{zhang2017citypersons} & 14.8         & -                          & -             \\
TLL+MRF \cite{TLL+MRF}                & 14.4         & 52.0                       & -             \\
RepLoss \cite{reploss}                & 13.2         & 56.9\textsuperscript{1}    & -             \\
OR-CNN \cite{OR-CNN}                  & 12.8         & 55.7\textsuperscript{1}    & -             \\
ALFNet \cite{ALF}                     & 12.0         & 51.9                       & -             \\
CSP \cite{liu2019CSP}                 & 11.0         & 49.3                       & -             \\
MGAN+ \cite{MGAN+}                    & 11.0         & 39.7                       & -             \\
KGSNet \cite{zhang2020kgsnet}         & 11.0         & 39.7                       & 36.2          \\
PRNet \cite{PRNet}                    & 10.8         & 42.0                       & -             \\
CSP+Detr \cite{csp+vit}               & 10.6         & 36.7                       & 36.4          \\
APD \cite{APD}                        & 9.7          & 41.3                       & 36.4          \\ \hline
CSP+MHAS (ours)                       & 10.7         & 40.8                       & 37.2          \\
CSP+Detr+MHAS (ours)                  & 10.3         & \textbf{35.8}              & \textbf{35.9} \\
APD+MHAS (ours)                       & \textbf{9.4} & 41.0                       & \textbf{35.9} \\ \hline
\end{tabular}%
\end{table}

\section{Discussion}
\label{sec: discussion}
In this paper, we explore the potential of the unique perspective property found in of the pedestrian datasets.
Compared to general object detection, the mean height of pedestrians is growing dramatically as they are closer to the viewpoint.
Building upon this characteristic, we propose the Mean Height Aided Suppression (MHAS) method in post-processing to improve the overall detection accuracy.
The existence score and mean height generators are developed to assist suppression.
Two approaches are proposed for mean height generation considering convenience and potential changes in eye levels.
The hyper-parameters in MHAS are carefully selected on a case-by-case basis through optimization on the pseudo-validation sets.
Importantly, the proposed methods are simple and can be universally applied to pedestrian detection tasks.
Comprehensive experiments show that both methods significantly improve the detection accuracy across various datasets and when applied to existing detectors. 
The combination of MHAS and APD achieves the best performance among the state-of-the-art.  
These findings highlight the efficacy and versatility of the proposed methods in advancing pedestrian detection.

\begin{figure}[!t]
\centering
\subfloat[]{\includegraphics[width=0.3\columnwidth]{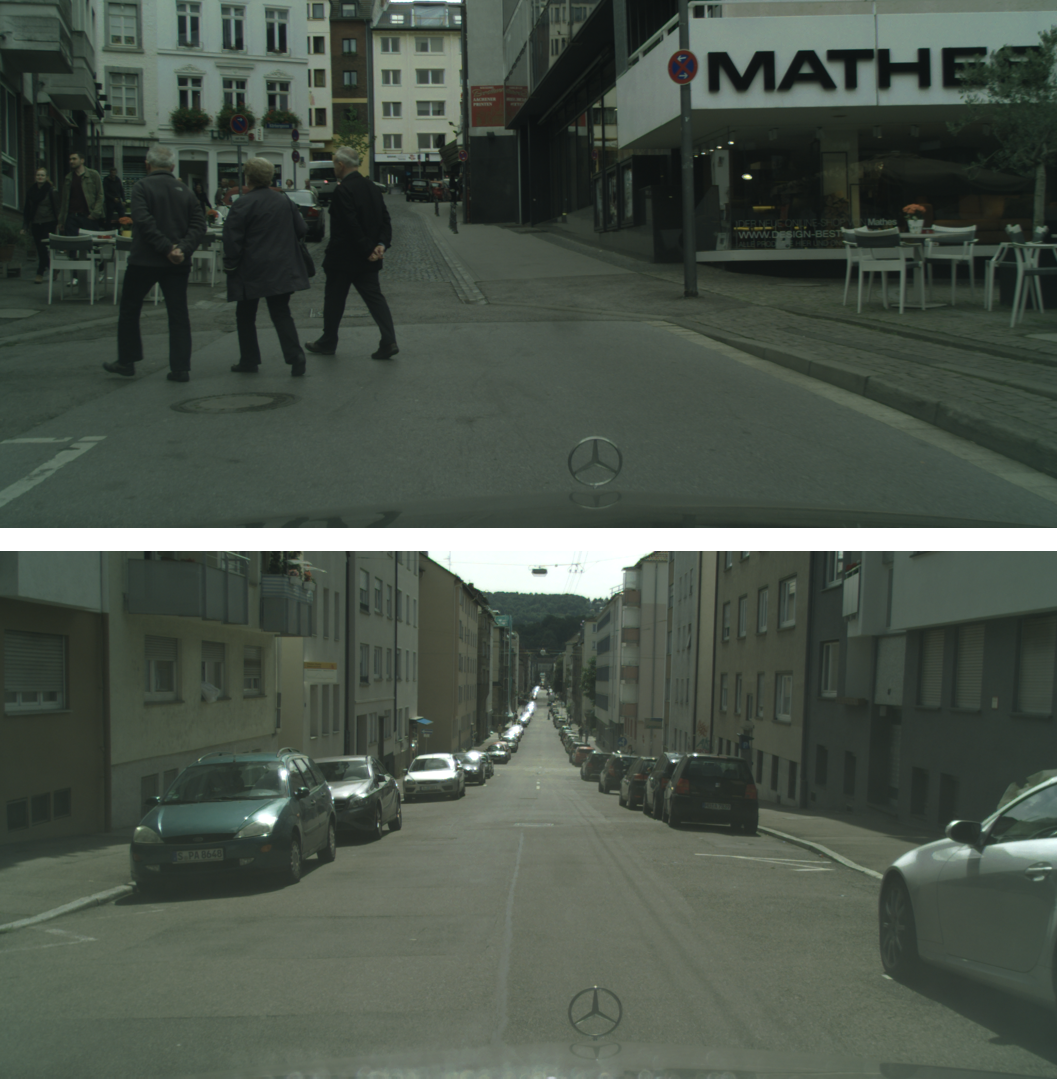}%
}
\hfil
\subfloat[]{\includegraphics[width=0.6\columnwidth]{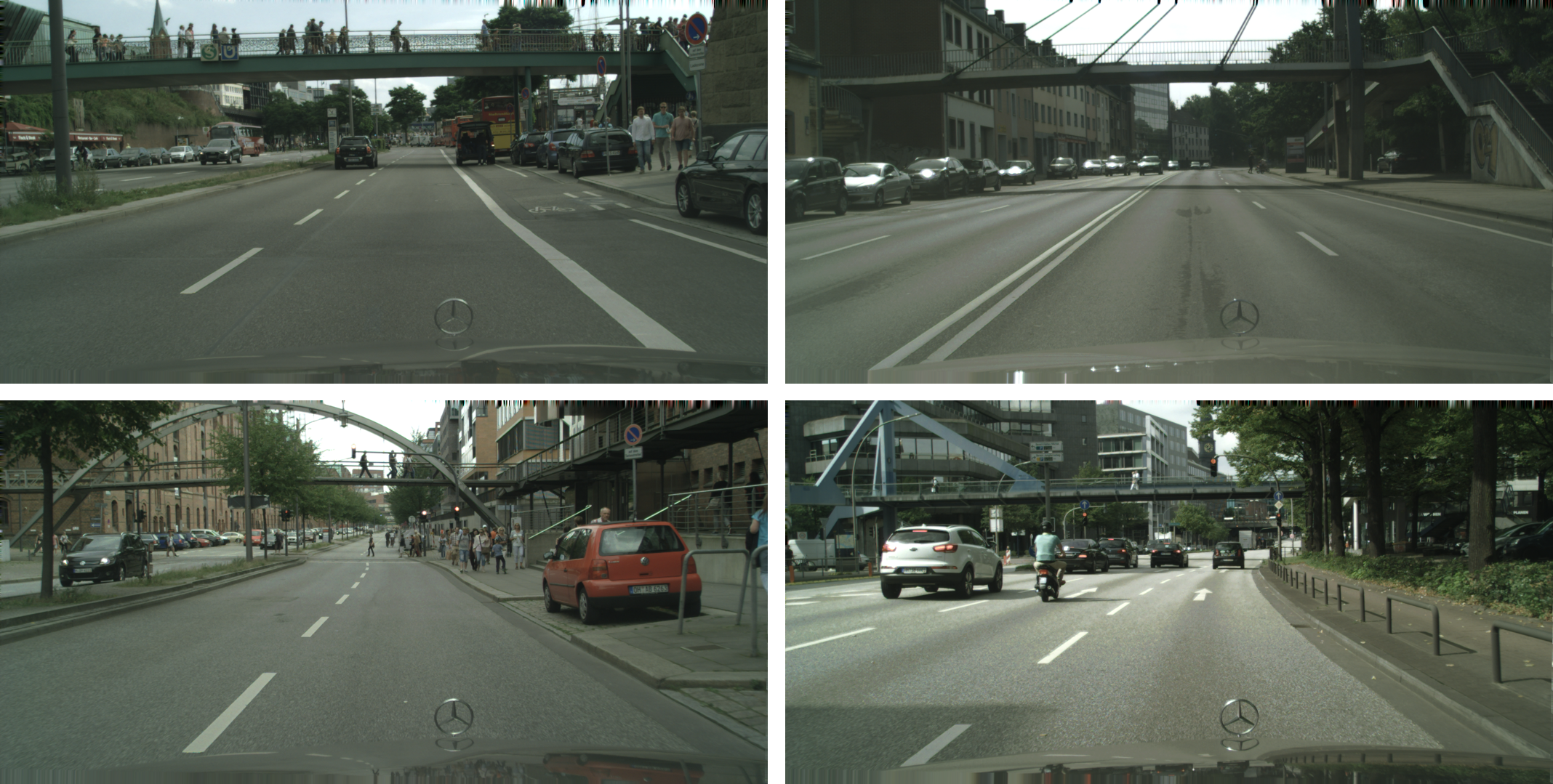}%
}
\caption{Exceptions of terrains that are beyond the consideration of the proposed algorithm.
(a) Samples of hills. 
(b) Samples of overpasses. Images are chosen from the Citypersons dataset.}
\label{fig: hills and overpass}
\end{figure}

The complete MHAS framework offers optimal performance. 
However, the existence score condition alone can also be used to achieve a certain extent of improvement faster, saving the time of mean height generation estimation.
This is particularly useful for cases where the mean heights are difficult to obtain with the two proposed approaches in the Mean Height Generator (MHG).
This is when dealing with scenarios where the eye levels of training images vary, and there is a scarcity of pedestrians in each image.
Note that the two-dimensional perspective is used. 
We assume that the datasets are obtained at the same altitude. 
Terrains like hills (Fig. \ref{fig: hills and overpass}(a)) or architectures like overpasses (Fig. \ref{fig: hills and overpass}(b)) are not considered in the algorithm design of MHAS in Section \ref{sec: mhas}.
In these cases, the mean height condition (Eq. \ref{eq:mh_filtering}) fails and may result in missed pedestrians.
To address this issue, there are two potential methods. 
The first method involves selecting hyperparameters that yield a high recall on the pseudo-validation set, as discussed in Section \ref{sec:mhas-choice of parameters}. 
This ensures a more comprehensive coverage of pedestrians, including those in challenging scenarios. 
The second method involves identifying problematic samples and applying MHAS only to the remaining samples that do not exhibit terrain or architectural complexities. 
By employing these methods, the overall performance of the pedestrian detection system can be improved, even in situations where mean height estimation faces limitations.

This paper leverages the under-explored mean height information as a unique prior knowledge of pedestrian detection tasks.
We believe such knowledge can hold a more significant role in task-specific object detection and may inspire advancements in the design of training, post-processing, and beyond. 
For example, the initialization of anchor size can be tailored to the mean height of pedestrians at different levels, thereby potentially accelerating convergence. 
Additionally, the network can be designed to possess shrinking receptive fields along the height dimension in high-level semantic features, an adaptation which can effectively suppress background noise and extract more accurate information for smaller targets, ultimately enhancing the detection performance.

\bibliographystyle{IEEEtran}
\bibliography{main}

\begin{thebibliography}{10}
\providecommand{\url}[1]{#1}
\csname url@samestyle\endcsname
\providecommand{\newblock}{\relax}
\providecommand{\bibinfo}[2]{#2}
\providecommand{\BIBentrySTDinterwordspacing}{\spaceskip=0pt\relax}
\providecommand{\BIBentryALTinterwordstretchfactor}{4}
\providecommand{\BIBentryALTinterwordspacing}{\spaceskip=\fontdimen2\font plus
\BIBentryALTinterwordstretchfactor\fontdimen3\font minus \fontdimen4\font\relax}
\providecommand{\BIBforeignlanguage}[2]{{%
\expandafter\ifx\csname l@#1\endcsname\relax
\typeout{** WARNING: IEEEtran.bst: No hyphenation pattern has been}%
\typeout{** loaded for the language `#1'. Using the pattern for}%
\typeout{** the default language instead.}%
\else
\language=\csname l@#1\endcsname
\fi
#2}}
\providecommand{\BIBdecl}{\relax}
\BIBdecl

\bibitem{RN62}
J.~Cao, Y.~Pang, J.~Xie, F.~S. Khan, and L.~Shao, ``From handcrafted to deep features for pedestrian detection: a survey,'' \emph{IEEE transactions on pattern analysis and machine intelligence}, 2021.

\bibitem{RN197}
W.~Li, X.~Zhu, and S.~Gong, ``Harmonious attention network for person re-identification,'' in \emph{Proceedings of the IEEE conference on computer vision and pattern recognition}, 2018, pp. 2285--2294.

\bibitem{RN198}
Y.~Li, J.~He, T.~Zhang, X.~Liu, Y.~Zhang, and F.~Wu, ``Diverse part discovery: Occluded person re-identification with part-aware transformer,'' in \emph{Proceedings of the IEEE/CVF Conference on Computer Vision and Pattern Recognition}, 2021, pp. 2898--2907.

\bibitem{RN199}
Y.~Yan, J.~Li, J.~Qin, S.~Bai, S.~Liao, L.~Liu, F.~Zhu, and L.~Shao, ``Anchor-free person search,'' in \emph{Proceedings of the IEEE/CVF Conference on Computer Vision and Pattern Recognition}, 2021, pp. 7690--7699.

\bibitem{RN201}
Y.~Li, S.~Zhang, Z.~Wang, S.~Yang, W.~Yang, S.-T. Xia, and E.~Zhou, ``Tokenpose: Learning keypoint tokens for human pose estimation,'' in \emph{Proceedings of the IEEE/CVF International Conference on Computer Vision}, 2021, pp. 11\,313--11\,322.

\bibitem{RN203}
Z.~Chen and X.~Huang, ``Pedestrian detection for autonomous vehicle using multi-spectral cameras,'' \emph{IEEE Transactions on Intelligent Vehicles}, vol.~4, no.~2, pp. 211--219, 2019.

\bibitem{RN204}
A.~Vobecky, M.~Uric{\'a}r, D.~Hurych, and R.~Skoviera, ``Advanced pedestrian dataset augmentation for autonomous driving,'' in \emph{Proceedings of the IEEE/CVF International Conference on Computer Vision Workshops}, 2019.

\bibitem{RN205}
Q.~Mou, L.~Wei, C.~Wang, D.~Luo, S.~He, J.~Zhang, H.~Xu, C.~Luo, and C.~Gao, ``Unsupervised domain-adaptive scene-specific pedestrian detection for static video surveillance,'' \emph{Pattern Recognition}, vol. 118, p. 108038, 2021.

\bibitem{dalal2005hog}
N.~Dalal and B.~Triggs, ``Histograms of oriented gradients for human detection,'' in \emph{2005 IEEE computer society conference on computer vision and pattern recognition (CVPR'05)}, vol.~1.\hskip 1em plus 0.5em minus 0.4em\relax Ieee, 2005, pp. 886--893.

\bibitem{dollar2014acf}
P.~Doll{\'a}r, R.~Appel, S.~Belongie, and P.~Perona, ``Fast feature pyramids for object detection,'' \emph{IEEE transactions on pattern analysis and machine intelligence}, vol.~36, no.~8, pp. 1532--1545, 2014.

\bibitem{cascade-rcnn}
Z.~Cai and N.~Vasconcelos, ``Cascade r-cnn: high quality object detection and instance segmentation,'' \emph{IEEE transactions on pattern analysis and machine intelligence}, vol.~43, no.~5, pp. 1483--1498, 2019.

\bibitem{faster-r-cnn}
S.~Ren, K.~He, R.~Girshick, and J.~Sun, ``Faster r-cnn: Towards real-time object detection with region proposal networks,'' \emph{Advances in neural information processing systems}, vol.~28, 2015.

\bibitem{liu2019CSP}
W.~Liu, S.~Liao, W.~Ren, W.~Hu, and Y.~Yu, ``High-level semantic feature detection: A new perspective for pedestrian detection,'' in \emph{Proceedings of the IEEE/CVF conference on computer vision and pattern recognition}, 2019, pp. 5187--5196.

\bibitem{ALF}
W.~Liu, S.~Liao, W.~Hu, X.~Liang, and X.~Chen, ``Learning efficient single-stage pedestrian detectors by asymptotic localization fitting,'' in \emph{Proceedings of the European Conference on Computer Vision (ECCV)}, 2018, pp. 618--634.

\bibitem{zhang2020kgsnet}
Y.~Zhang, Y.~Bai, M.~Ding, S.~Xu, and B.~Ghanem, ``Kgsnet: key-point-guided super-resolution network for pedestrian detection in the wild,'' \emph{IEEE transactions on neural networks and learning systems}, vol.~32, no.~5, pp. 2251--2265, 2020.

\bibitem{mgan2019}
Y.~Pang, J.~Xie, M.~H. Khan, R.~M. Anwer, F.~S. Khan, and L.~Shao, ``Mask-guided attention network for occluded pedestrian detection,'' in \emph{Proceedings of the IEEE/CVF international conference on computer vision}, 2019, pp. 4967--4975.

\bibitem{MGAN+}
J.~Xie, Y.~Pang, M.~H. Khan, R.~M. Anwer, F.~S. Khan, and L.~Shao, ``Mask-guided attention network and occlusion-sensitive hard example mining for occluded pedestrian detection,'' \emph{IEEE transactions on image processing}, vol.~30, pp. 3872--3884, 2020.

\bibitem{chi2020pedhunter}
C.~Chi, S.~Zhang, J.~Xing, Z.~Lei, S.~Z. Li, and X.~Zou, ``Pedhunter: Occlusion robust pedestrian detector in crowded scenes,'' in \emph{Proceedings of the AAAI Conference on Artificial Intelligence}, vol.~34, no.~07, 2020, pp. 10\,639--10\,646.

\bibitem{zhang2017citypersons}
S.~Zhang, R.~Benenson, and B.~Schiele, ``Citypersons: A diverse dataset for pedestrian detection,'' in \emph{Proceedings of the IEEE conference on computer vision and pattern recognition}, 2017, pp. 3213--3221.

\bibitem{dollar2011pedestrian}
P.~Dollar, C.~Wojek, B.~Schiele, and P.~Perona, ``Pedestrian detection: An evaluation of the state of the art,'' \emph{IEEE transactions on pattern analysis and machine intelligence}, vol.~34, no.~4, pp. 743--761, 2011.

\bibitem{coco}
T.-Y. Lin, M.~Maire, S.~Belongie, J.~Hays, P.~Perona, D.~Ramanan, P.~Doll{\'a}r, and C.~L. Zitnick, ``Microsoft coco: Common objects in context,'' in \emph{European conference on computer vision}.\hskip 1em plus 0.5em minus 0.4em\relax Springer, 2014, pp. 740--755.

\bibitem{csp+vit}
J.~Yuan, P.~Barmpoutis, and T.~Stathaki, ``Multi-scale deformable transformer encoder based single-stage pedestrian detection,'' in \emph{2022 IEEE International Conference on Image Processing (ICIP)}.\hskip 1em plus 0.5em minus 0.4em\relax IEEE, 2022, pp. 2906--2910.

\bibitem{reploss}
X.~Wang, T.~Xiao, Y.~Jiang, S.~Shao, J.~Sun, and C.~Shen, ``Repulsion loss: Detecting pedestrians in a crowd,'' in \emph{Proceedings of the IEEE Conference on Computer Vision and Pattern Recognition}, 2018, pp. 7774--7783.

\bibitem{leibe2008coupled}
B.~Leibe, K.~Schindler, N.~Cornelis, and L.~Van~Gool, ``Coupled object detection and tracking from static cameras and moving vehicles,'' \emph{IEEE transactions on pattern analysis and machine intelligence}, vol.~30, no.~10, pp. 1683--1698, 2008.

\bibitem{hoiem2008putting}
D.~Hoiem, A.~A. Efros, and M.~Hebert, ``Putting objects in perspective,'' \emph{International Journal of Computer Vision}, vol.~80, pp. 3--15, 2008.

\bibitem{chen2016monocular}
X.~Chen, K.~Kundu, Z.~Zhang, H.~Ma, S.~Fidler, and R.~Urtasun, ``Monocular 3d object detection for autonomous driving,'' in \emph{Proceedings of the IEEE conference on computer vision and pattern recognition}, 2016, pp. 2147--2156.

\bibitem{liang2018MultiSensor3D}
M.~Liang, B.~Yang, S.~Wang, and R.~Urtasun, ``Deep continuous fusion for multi-sensor 3d object detection,'' in \emph{Proceedings of the European conference on computer vision (ECCV)}, 2018, pp. 641--656.

\bibitem{huang2019perspectivenet}
S.~Huang, Y.~Chen, T.~Yuan, S.~Qi, Y.~Zhu, and S.-C. Zhu, ``Perspectivenet: 3d object detection from a single rgb image via perspective points,'' \emph{Advances in neural information processing systems}, vol.~32, 2019.

\bibitem{kocur2020perspectiveTransformation}
V.~Kocur and M.~Ft{\'a}{\v{c}}nik, ``Detection of 3d bounding boxes of vehicles using perspective transformation for accurate speed measurement,'' \emph{Machine Vision and Applications}, vol.~31, no. 7-8, p.~62, 2020.

\bibitem{wang2022probabilistic}
T.~Wang, Z.~Xinge, J.~Pang, and D.~Lin, ``Probabilistic and geometric depth: Detecting objects in perspective,'' in \emph{Conference on Robot Learning}.\hskip 1em plus 0.5em minus 0.4em\relax PMLR, 2022, pp. 1475--1485.

\bibitem{he2016resnet}
K.~He, X.~Zhang, S.~Ren, and J.~Sun, ``Deep residual learning for image recognition,'' in \emph{Proceedings of the IEEE conference on computer vision and pattern recognition}, 2016, pp. 770--778.

\bibitem{zhang2016far}
S.~Zhang, R.~Benenson, M.~Omran, J.~Hosang, and B.~Schiele, ``How far are we from solving pedestrian detection?'' in \emph{Proceedings of the IEEE conference on computer vision and pattern recognition}, 2016, pp. 1259--1267.

\bibitem{APD}
J.~Zhang, L.~Lin, J.~Zhu, Y.~Li, Y.-c. Chen, Y.~Hu, and S.~C. Hoi, ``Attribute-aware pedestrian detection in a crowd,'' \emph{IEEE Transactions on Multimedia}, vol.~23, pp. 3085--3097, 2020.

\bibitem{RPN+BF}
L.~Zhang, L.~Lin, X.~Liang, and K.~He, ``Is faster r-cnn doing well for pedestrian detection?'' in \emph{European conference on computer vision}.\hskip 1em plus 0.5em minus 0.4em\relax Springer, 2016, pp. 443--457.

\bibitem{jointDet}
C.~Chi, S.~Zhang, J.~Xing, Z.~Lei, S.~Z. Li, and X.~Zou, ``Relational learning for joint head and human detection,'' in \emph{Proceedings of the AAAI Conference on Artificial Intelligence}, vol.~34, no.~07, 2020, pp. 10\,647--10\,654.

\bibitem{TLL+MRF}
T.~Song, L.~Sun, D.~Xie, H.~Sun, and S.~Pu, ``Small-scale pedestrian detection based on topological line localization and temporal feature aggregation,'' in \emph{Proceedings of the European Conference on Computer Vision (ECCV)}, 2018, pp. 536--551.

\bibitem{OR-CNN}
S.~Zhang, L.~Wen, X.~Bian, Z.~Lei, and S.~Z. Li, ``Occlusion-aware r-cnn: detecting pedestrians in a crowd,'' in \emph{Proceedings of the European Conference on Computer Vision (ECCV)}, 2018, pp. 637--653.

\bibitem{PRNet}
X.~Song, K.~Zhao, W.-S. Chu, H.~Zhang, and J.~Guo, ``Progressive refinement network for occluded pedestrian detection,'' in \emph{European Conference on Computer Vision}.\hskip 1em plus 0.5em minus 0.4em\relax Springer, 2020, pp. 32--48.

\end{thebibliography}

\end{document}